\documentclass{article}

\usepackage[utf8]{inputenc}
\usepackage[T1]{fontenc}
\usepackage{url}
\usepackage{booktabs}
\usepackage{amsfonts}
\usepackage{nicefrac}
\usepackage[dvipsnames]{xcolor}
\usepackage{hyperref}
\usepackage{color}
\usepackage{epsfig}

\usepackage{amsthm}
\usepackage{amsmath}
\usepackage{float}
\usepackage{graphicx}
\usepackage{wrapfig}
\usepackage[capitalise]{cleveref}
\usepackage{makecell}
\usepackage{caption}
\usepackage{subcaption}
\usepackage{bm}
\usepackage{multirow}
\usepackage{microtype}
\usepackage{mathtools}
\usepackage{tikz}
\usepackage{gensymb}
\usepackage{bbm}
\usepackage{chngcntr}
\usepackage{amssymb}
\usepackage{enumitem}
\usepackage{placeins}
\usepackage{etoolbox}
\usepackage{tabularx,colortbl}
\usepackage{tgtermes}

\usepackage[accepted]{icml2020}

\newcommand{\MyMapTemplatePrefixc}[4]{\expandafter#1\csname#3#4\endcsname{#2{#4}}} 
\forcsvlist{\MyMapTemplatePrefixc {\def} {\mathcal}{c}} {A,B,C,D,E,F,G,H,I,J,K,L,M,N,O,P,Q,R,S,T,U,V,W,X,Y,Z}  
\newcommand{\MyMapTemplatePrefixtb}[5]{\expandafter#1\csname#4#5\endcsname{#2{#3{#5}}}} 
\forcsvlist{\MyMapTemplatePrefixtb {\def} {\tilde}{\mathbf}{t}} {A,B,C,D,E,F,G,H,I,J,K,L,M,N,O,P,Q,R,S,T,U,V,W,X,Y,Z}  
\forcsvlist{\MyMapTemplatePrefixtb {\def} {\tilde}{\mathbf}{t}} {0,1,a,b,c,d,e,f,g,h,i,j,k,l,m,n,o,p,q,r,s,u,v,w,x,y,z}  
\makeatletter
\newcommand\footnoteref[1]{\protected@xdef\@thefnmark{\ref{#1}}\@footnotemark}
\makeatother
\newcommand{\MyMapTemplateNoPrefix}[3]{\expandafter#1\csname#3\endcsname{#2{#3}}}
\forcsvlist{\MyMapTemplatePrefixc {\def} {\mathbf}{mb}} {0,1,a,b,c,d,e,f,g,h,i,j,k,l,m,n,o,p,q,r,s,u,v,w,x,y,z}  
\forcsvlist{\MyMapTemplatePrefixc {\def} {\mathbf}{mb}} {A,B,C,D,E,F,G,H,I,J,K,L,M,N,O,P,Q,R,S,T,U,V,W,X,Y,Z}

\forcsvlist{\MyMapTemplatePrefixc {\def} {\mathrm}{rm}} {0,1,a,b,c,d,e,f,g,h,i,j,k,l,m,n,o,p,q,r,s,u,v,w,x,y,z}  
\forcsvlist{\MyMapTemplatePrefixc {\def} {\mathrm}{rm}} {A,B,C,D,E,F,G,H,I,J,K,L,M,N,O,P,Q,R,S,T,U,V,W,X,Y,Z}    

\newcommand{\specialcell}[2][c]{%
  \begin{tabular}[#1]{@{}l@{}}#2\end{tabular}}

\newcommand{\ie}{\textit{i}.\textit{e}., }
\newcommand{\eg}{\textit{e}.\textit{g}., }

\crefformat{section}{\S#2#1#3}
\crefformat{subsection}{\S#2#1#3}
\crefformat{subsubsection}{\S#2#1#3}

\definecolor{lightgray}{gray}{.92}

\newcommand{\cgr}{\cellcolor{lightgray}}
\newcommand{\bt}{\textbf}

\definecolor{pinkish}{HTML}{fc41a2}
\newcommand{\MYhref}[3][pinkish]{\href{#2}{\color{#1}{#3}}}%

\icmltitlerunning{Spatial Contrastive Learning for Few-Shot Classification}

\begin{document}

\twocolumn[\icmltitle{Spatial Contrastive Learning for Few-Shot Classification}

\author{
  Yassine Ouali\hspace{3pc}
  Céline Hudelot\hspace{3pc}
  Myriam Tami\\
  Université Paris-Saclay, CentraleSupélec, MICS, 91190, Gif-sur-Yvette, France.\\
  {\tt\small \{yassine.ouali,celine.hudelot,myriam.tami\}@centralesupelec.fr}
}

\icmlsetsymbol{equal}{*}

\begin{icmlauthorlist}
\icmlauthor{Yassine Ouali}{}
\icmlauthor{Céline Hudelot}{}
\icmlauthor{Myriam Tami}{} \\
Université Paris-Saclay, CentraleSupélec, MICS, 91190, Gif-sur-Yvette, France.\\
{\tt\small \{yassine.ouali,celine.hudelot,myriam.tami\}@centralesupelec.fr}
\end{icmlauthorlist}

\icmlkeywords{Contrastive Learning, Few-shot Classification, Deep Learning}

\vskip 0.5in
]

\begin{abstract}

In this paper, we explore contrastive learning %
for few-shot classification, in which %
we propose to use it as an additional auxiliary training objective acting as a data-dependent
regularizer to promote more general and transferable features.
In particular, we present a novel attention-based spatial contrastive
objective to learn locally discriminative and class-agnostic features.
As a result, our approach overcomes some of the limitations of the
cross-entropy loss, such as its excessive discrimination towards seen
classes, which reduces the transferability of features to unseen classes.
With extensive experiments, we show that the proposed method outperforms
state-of-the-art approaches, confirming the importance of learning good
and transferable embeddings for few-shot learning
\\ Code: \MYhref{https://github.com/yassouali/SCL}{https://github.com/yassouali/SCL}.
\end{abstract}

\section{Introduction}
\label{sec:intro}

Few-shot learning \cite{lake2011one} has emerged as an alternative to
supervised learning to simulate more realistic settings that mimic human capabilities,
and in particular, it consists of reproducing the learner's ability
to rapidly and efficiently adapt to novel tasks.
In this paper, we tackle the problem of few-shot image classification, which aims
to equip a learner with the ability to learn novel visual
concepts and recognize unseen classes with limited supervision.

A popular paradigm to solve this problem is meta-learning \cite{thrun1998lifelong,naik1992meta}
consisting of two disjoint stages, meta-training and meta-testing.
During meta-training, the goal is to acquire transferable knowledge from a set of
tasks sampled from the meta-training tasks
so that the learner is equipped with the ability to adapt to novel tasks quickly.
This fast adaptability to unseen classes
is evaluated at test time by the average test accuracy over several meta-testing tasks.
Such transferable knowledge can be acquired from the meta-training tasks with
optimization-based methods \cite{ravi2016optimization, finn2017model}
or metric-based methods \cite{vinyals2016matching, snell2017prototypical, sung2018learning}.

\begin{figure}[t]
  \centering
  \includegraphics[width=0.85\linewidth]{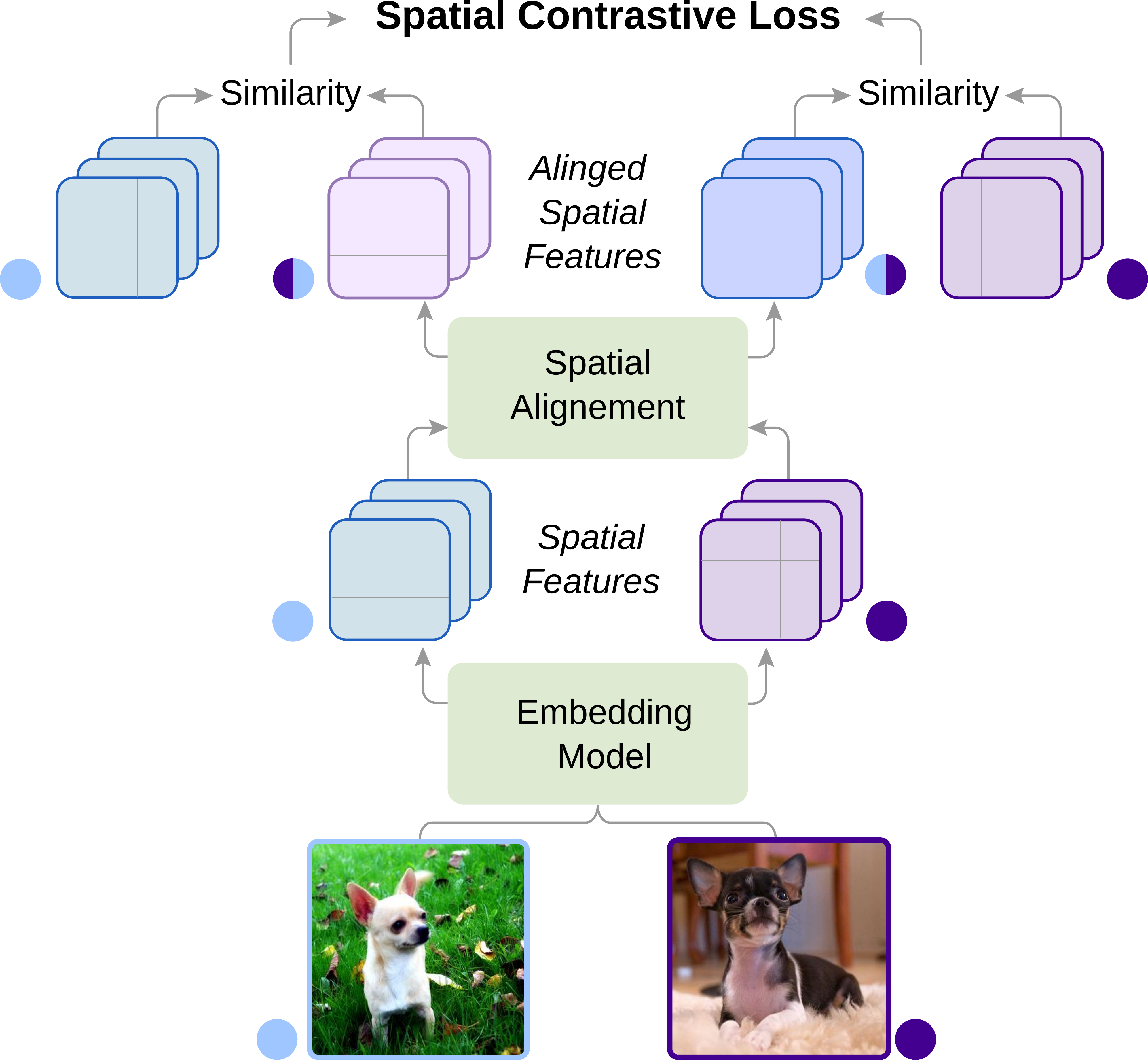}
  \vspace{-0.05in}
  \caption{{Spatial Contrastive Learning (SCL).}
    To learn more locally class-independent discriminative features,
    we propose to measure the similarity between a given pair of samples using 
    their spatial features as opposed to their global features.
    We first apply an attention-based
    alignment, aligning each input with respect to the other. Then,
    we measure the one-to-one spatial similarities and compute the Spatial Contrastive (SC) loss.}
  \label{fig:overview}
  \vspace{-0.25in}
\end{figure}

Recently, a growing line of works
\cite{chen2019closer,dhillon2019baseline,tian2020rethinking} show
that learning good representations results in fast adaptability at test time,
suggesting that feature reuse \cite{raghu2019rapid}
plays a more important role in few-shot classification than the meta-learning aspect
of existing algorithms. Such methods consider an extremely simple transfer learning baseline, in which
the model is first pre-trained using the standard cross-entropy (CE) loss on
the meta-training set. Then, at test time, a linear classifier is trained on the
meta-testing set on top of the pre-trained model. The pre-trained model can either be fine-tuned
\cite{dhillon2019baseline,afrasiyabi2019associative} together with the classifier,
or fixed and used as a feature extractor \cite{chen2019closer,tian2020rethinking}.
While promising, we argue that using the CE loss during the pre-training stage
hinders the quality of the learned representations since the model
only acquires the necessary knowledge to solve the classification task over seen
classes at train time. As a result, the learned visual features are excessively
discriminative against the training classes, rendering them sub-optimal for test
time classification tasks constructed from an arbitrary set of unseen and novel classes.

To alleviate these limitations, we propose to leverage contrastive representation learning \cite{wu2018unsupervised, he2020momentum, chen2020simple} as
an auxiliary objective, where instead of only mapping the inputs to fixed targets,
we also optimize the features, pulling together semantically similar (\ie positive) samples in the embedding space while pushing apart dissimilar (\ie negative) samples.
By integrating the contrastive loss into the learning objective, we give
rise to discriminative representations between dissimilar instances
while maintaining an invariance towards visual similarities.
Subsequently, the learned representations are more transferable and capture more prevalent patterns outside of the seen classes. Additionally, by combining both losses, we leverage the stability of the CE loss and its effectiveness on small datasets and small batch sizes, while taking benefit of the contrastive
loss as a data-dependent regularizer promoting more general-purpose embeddings.
Additionally, by combining both losses, we leverage the stability of the CE loss and its effectiveness on small datasets
and small batch sizes, in addition to taking benefit of the contrastive loss as a data-dependent regularizer promoting more general-purpose embeddings.

Specifically, we propose a novel attention-based spatial contrastive loss (see \cref{fig:overview})
as the auxiliary objective to further promote class-agnostic visual features and avoid suppressing local discriminative patterns. It consists of measuring the local similarity between the spatial features of a given pair of samples after an attention-based spatial alignment mechanism, instead of the
global features (\ie avg. pooled spatial features) used in the standard contrastive loss.
We also adopt the supervised formulation \cite{khosla2020supervised} of
the contrastive loss to leverage the provided label information
when constructing the positive and negative samples.

However, directly optimizing the features and promoting the formation of
clusters of similar instances in the embedding space might result in extremely disentangled representations.
Such an outcome can be undesirable for few-shot learning, where the testing tasks can be notably different
from the tasks encountered during training, \eg training on generic categories, and testing on fine-grained sub-categories. 
To solve this, we propose contrastive distillation to reduce the compactness of the features in the embedding space and provide additional refinement of the representations.\\

\par \noindent \textbf{Contributions.} To summarize, our contributions are:
\begin{itemize}
\item We explore contrastive learning as an auxiliary pre-training objective
to learn more transferable features. 

\item We propose a novel Spatial Contrastive (SC) loss with an
attention-based alignment mechanism to spatially compare a pair of features,
further promoting class-independent discriminative patterns.

\item We employ contrastive distillation to avoid excessive disentanglement of the learned embeddings
and improve the performances.

\item We demonstrate the effectiveness of the proposed method with extensive
experiments on standard and cross-domain few-shot classification
benchmarks, achieving state-of-the-art performances.

\item We show the universality of the proposed method by applying it to a standard metric learning approach, resulting
in a notable performance boost.
\end{itemize}

\begin{figure*}[t]
  \centering
  \includegraphics[width=0.95\linewidth]{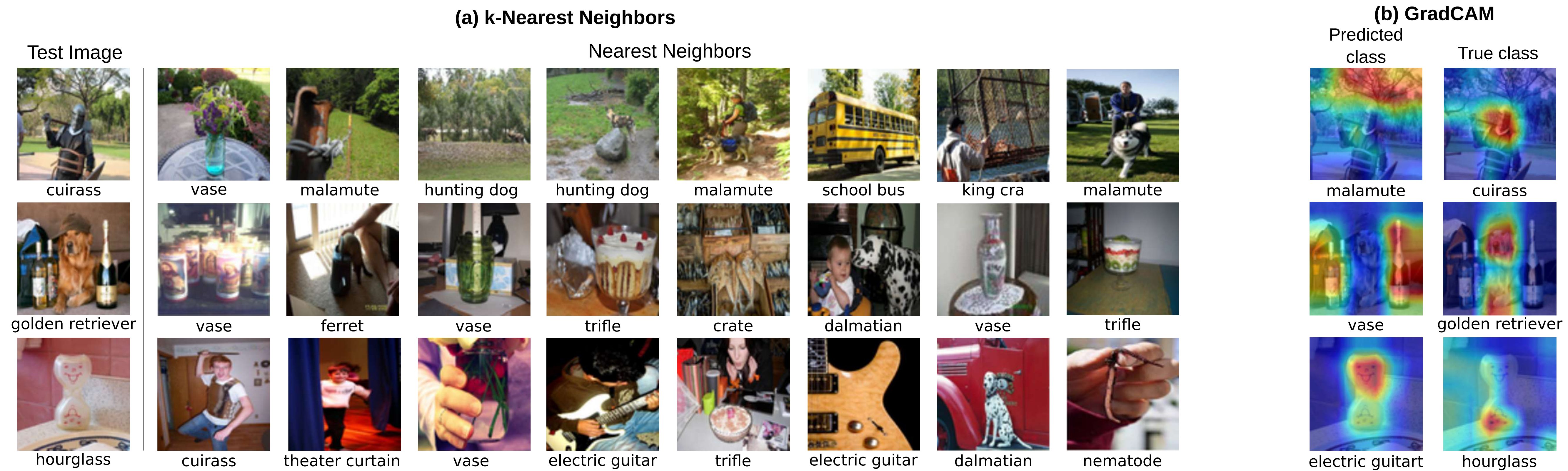}
  \vspace{-0.1in}
  \caption{{Analysis of the Learned Representations.} (a) $k$-Nearest Neighbors Analysis.
  For a given test image from \textit{mini}-ImageNet dataset,
  we compute the nearest neighbors in the embedding space on the test set, and we observe that they are
  semantically dissimilar. This suggests that the learned embeddings are excessively discriminative
  towards features used to solve the training classification tasks
  (\eg the beer bottles in the second test image),
  which are not useful to recognize the novel classes at test time.
  (b) GradCAM results. To obtain the class activation maps (CAMs) explaining 
  such an outcome, we train a linear classifier on the whole test set on
  top of the frozen embedding model and compute the CAMs. We see that the dominant
  discriminative features are not the ones useful for test-time classification.}
  \label{fig:features_analysis}
  \vspace{-0.15in}
\end{figure*}

\vspace{-0.2in}
\section{Preliminaries}
\label{sec:}

Following a similar notation as \cite{tian2020rethinking,lee2019meta}, we start by introducing
the meta-learning formulation and the standard transfer learning
baseline of \cite{tian2020rethinking} in \cref{subsec:problem} and \cref{subsec:TL}. Then, in \cref{subsec:quality_of_features},
we analyze the quality of the learned features in such a setting, motivating
the need for an alternative pre-training objective in order to learn more transferable embeddings.

\subsection{Problem Definition}
\label{subsec:problem}

Few-shot classification usually involves a meta-training set $\cT$ and a meta-testing set
$\cS$ with disjoint label spaces.
The meta-training set discerns \textit{seen} classes, while the meta-testing set discerns
novel and \textit{unseen} classes. Each one of the meta sets consists of
a number of classification tasks where each task
describes a pair of training (\ie support) and testing (\ie query) sets with few examples, \ie 
$\cT=\{(\cD_{t}^{\mathrm{train}}, \cD_{t}^{\mathrm{test}})\}_{t=1}^{T}$ and 
$\cS=\{(\cD_{q}^{\mathrm{train}}, \cD_{q}^{\mathrm{test}})\}_{q=1}^{Q}$, with each
dataset containing pairs of images $\mbx$ and their ground-truth labels $y$.

The goal of few-shot classification is to learn a classifier $f_\theta$ parametrized by $\theta$
capable of exploiting the few training examples provided by the dataset $\cD^{\mathrm{train}}$
to correctly predict the labels of the test examples from $\cD^{\mathrm{test}}$
for a given task. However, 
given the high dimensionality of the inputs and the limited number of training examples,
the classifier $f_\theta$ suffers from high variance.
As such, the training and testing inputs are replaced with their corresponding features, which are produced by an embedding model $f_\phi$ parametrized by $\phi$ and then used as inputs to the classifier $f_\theta$.

To this end, the objective of meta-training algorithms is to learn a good embedding model
$f_\phi$ so that the average test error of the classifier $f_\theta$ is minimized.
This usually involves two stages: first, a meta-training stage inferring
the parameters $\phi$ of the embedding model using the meta-training
set $\cT$, followed by a meta-testing stage evaluating the embedding model's
performance on meta-testing set $\cS$.

\subsection{Transfer Learning Baseline}
\label{subsec:TL}

In this work, we consider the simple transfer learning baseline of \cite{tian2020rethinking},
in which the embedding model $f_\phi$ is first pre-trained on the merged tasks
from the meta-training set using the CE loss.
Then, the model is carried over to the meta-testing stage and fixed during evaluation.

Concretely, we start by merging all 
the  meta-training tasks $\cD_{t}^{\mathrm{train}}$ from $\cT$ into a 
single training set $\cD^{\mathrm{new}}$ of seen classes:
\begin{equation}
\cD^{\mathrm{new}}
=\cup \{\cD_{1}^{\mathrm{train}}, \ldots, \cD_{t}^{\mathrm{train}}, \ldots, \cD_{T}^{\mathrm{train}}\}.
\end{equation}
Then, during the meta-training stage, the embedding model $f_\phi$ can be pre-trained on the resulting
set of seen classes using the standard CE loss $L_\mathrm{CE}$:
\begin{equation}
\phi = \underset{\phi}{\arg\min}\ L_\mathrm{CE} (\cD^\mathrm{new} ; \phi).
\label{eq:CE}
\end{equation}
The pre-trained model $f_\phi$ is then fixed (\ie no fine-tuning is performed)
and leveraged as a feature extractor during the meta-testing stage.
For a given task $(\cD_{q}^{\mathrm{train}}, \cD_{q}^{\mathrm{test}})$ sampled from
$\cS$, a linear classifier $f_\theta$ is first trained on top of the extracted features to recognize
the unseen classes using the training dataset $\cD_{q}^{\mathrm{train}}$:
\begin{equation}
\theta = \underset{\theta}{\arg\min}\ L_\mathrm{CE} (\cD_{q}^{\mathrm{train}}; \theta, \phi) + \cR(\theta),
\end{equation}
where $\cR$ is a regularization term, and the parameters
$\theta = \{\mbW, \mbb\}$ consist of weight and bias terms, respectively.
The predictor $f_\theta$ can then be used on the features of the test dataset $\cD_{q}^{\mathrm{test}}$
to obtain the class predictions and evaluate $f_\phi$.

\subsection{Analysis of the Learned Representations}
\label{subsec:quality_of_features}

Although the baseline of \cref{subsec:TL} delivers impressive results, we hypothesis that the usage of the CE loss during the meta-training stage can hinder the performances. 
Our intuition is that the learned representations lack general discriminative visual features since the CE loss induces %
embeddings tailored for solving the classification task over the seen classes. As a results, their transferability to novel domains with unseen classes is reduced, and especially if the domain gap between
the training and testing stages is significant.

To empirically validate such a hypothesis, we conduct a $k$-nearest neighbor search
\cite{faiss} on the learned embedding space.
First, we train a model with the CE loss on the meta-training set of
\textit{mini}-ImageNet \cite{vinyals2016matching} as in \cref{eq:CE}.
Then, for a given test image, we search for 
its neighbors from the meta-testing set.
The results are shown in \cref{fig:features_analysis}.
For a fast test-time adaptation of the predictor $f_\theta$, the desired outcome is to
have visually and semantically similar images adjacent
in the embedding space. However, we observe that the
neighboring images are semantically dissimilar. Using Grad-CAM \cite{selvaraju2017grad}, we notice that dominant
discriminative features acquired during training might not be useful
for discriminating between unseen classes at test time.
In the case of \textit{mini}-ImageNet, this observation is reinforced by
the fact that the meta-training and meta-testing sets are closely related, in which
better transferability of the learned features in expected when compared to other benchmarks.
We note that similar behavior was also observed by \cite{doersch2020crosstransformers}
for metric-learning based approaches.

\begin{figure}[t]
  \centering
  \includegraphics[width=0.9\linewidth]{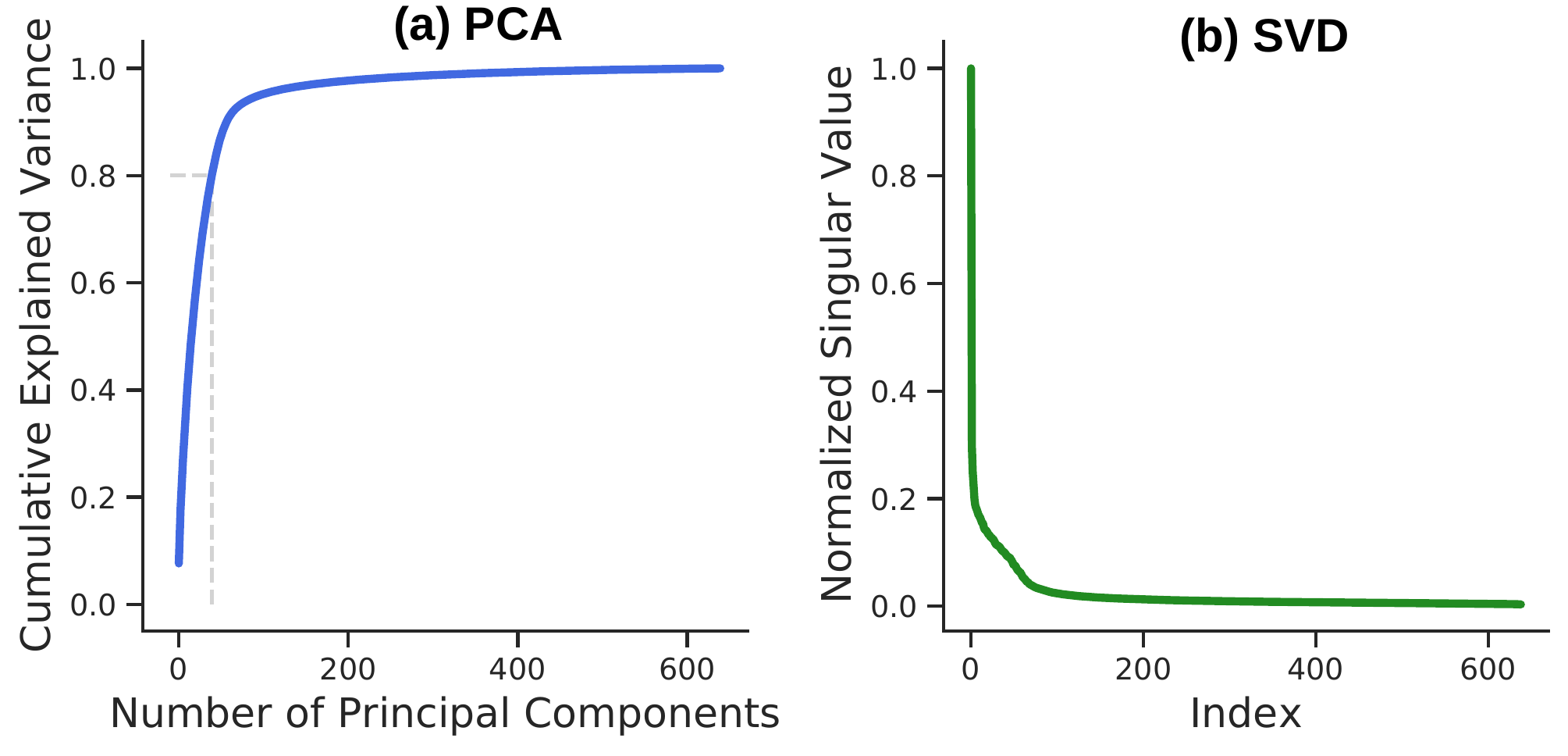}
  \vspace{-0.1in}
  \caption{{Spectral Analysis.} Results of the spectral analysis
  on the embedding matrix. The plot (a) shows the explained
  cumulative variance of the learned features as the number of principal components used.
  We observe that $80\%$ of the variance can be explained with only $30$ components,
  indicating that embeddings lie in a lower dimensional space and are discriminative
  towards a small number of visual structures. Similarly, by computing the singular values
  of the embedding matrix, we see in (b) that the first singular values dominate
  the rest, indicating the same behavior.}
  \label{fig:spectral_analysis}
  \vspace{-0.15in}
\end{figure}

To further investigate this behavior, we conduct a spectral analysis of the learned features. 
As shown in \cref{fig:spectral_analysis},
we inspect the variance explained by
a varying number of principal components and notice that almost all of the variance can
be captured with a limited number of components,
indicating that the CE loss only preserves the minimal amount 
of information required to solve the classification task.
Similarity, by applying singular value decomposition to compute the eigen values of the feature matrix,
we observe that the maximal singular values are significantly larger than the remaining ones,
diminishing the amount of informative signal that can be captured.

\section{Methodology}
\label{sec:methods}

Based on the observations presented in \cref{subsec:quality_of_features}, in this section,
we explore an alternative pre-training objective in order to learn a more transferable
embedding model $f_\phi$. First, in \cref{subsec:GCL}, we present the standard supervised
contrastive loss. Then, in \cref{subsec:SCL},
we introduce a novel spatial contrastive learning objective followed by
the pre-training objective in \cref{subsec:total_loss}. Finally, an optional
contrastive distillation step in \cref{subsec:regulatization}.

\subsection{Contrastive Learning}
\label{subsec:GCL}

We explore contrastive learning as an auxiliary pre-training objective
to learn general-purpose visual embeddings capturing discriminative features usable outside of the meta-training set. It thus facilitate the test time recognition of unseen classes.
Specifically, given that in a few-shot classification setting we are provided with the class labels,
we examine the usage of the supervised formulation \cite{khosla2020supervised} of the contrastive loss
which leverages the label information to construct the positive and negative samples.

Formally, let $f_\phi$ be an embedding model mapping the inputs $\mbx$ to \textit{spatial}
features $\mbz^\rms \in \mathbb{R}^{HW \times d}$,
followed by an average pooling operation to obtain the \textit{global}
features $\mbz^\rmg \in \mathbb{R}^{d}$, which are then
mapped into a lower dimensional space using a projection head $p$, \ie
$\mbf = p(\mbz^\rmg)$ with $\mbf \in \mathbb{R}^{d^\prime}$, and let a global similarity function $\mathrm{sim}_{\mathrm{g}}$
be denoted as the cosine similarity between a pair of projected global features $\mbf_i$ and $\mbf_j$
(\ie dot product between the $\ell_2$ normalized features).
First, we sample a batch of $N$ pairs
of images and labels from the merged meta-training set $\cD^\mathrm{new}$ and
augment each example in the batch, resulting in $2N$ data points.
Then, the supervised contrastive loss \cite{khosla2020supervised}, referred to as the Global Contrastive (GC) loss, can be computed as follows:
\begin{gather}
\label{eq:GCL}
L_{\rm{GC}} = \sum_{i=1}^{2N} \frac{1}{2 N_{y_i} - 1} \sum_{j=1}^{2N} \mathbbm{1}_{i \neq j} \cdot \mathbbm{1}_{y_i=y_j} \cdot \ell_{ij}\ , \\
\text{where}\ \ \ell_{ij} = - \log \frac{\exp(\mathrm{sim}_{\mathrm{g}}(\mbf_i,\mbf_j) / \tau)}
{\sum_{k=1}^{2N} \mathbbm{1}_{i \neq k} \cdot \exp (\mathrm{sim}_{\mathrm{g}}(\mbf_i,\mbf_k) / \tau)}\ ,  \notag 
\end{gather}
with $\mathbbm{1}_{\mathrm{cond}} \in\{0,1\}$ as an indicator function evaluating to $1$ iff $\mathrm{cond}$ is satisfied,
$N_{y_i}$ as the total number of images with the same label $y_i$, and $\tau$ as a scalar temperature parameter.
By using the GC loss of \cref{eq:GCL}
as an additional pre-training objective with the CE loss, we push the embedding model $f_\phi$ to learn the visual similarities
between instances of the same class, instead of only maintaining the useful features for the classification task over the seen classes, which results in more useful and transferable embeddings.

\subsection{Spatial Contrastive Learning}
\label{subsec:SCL}

\begin{figure}[t]
  \centering
  \includegraphics[width=0.9\linewidth]{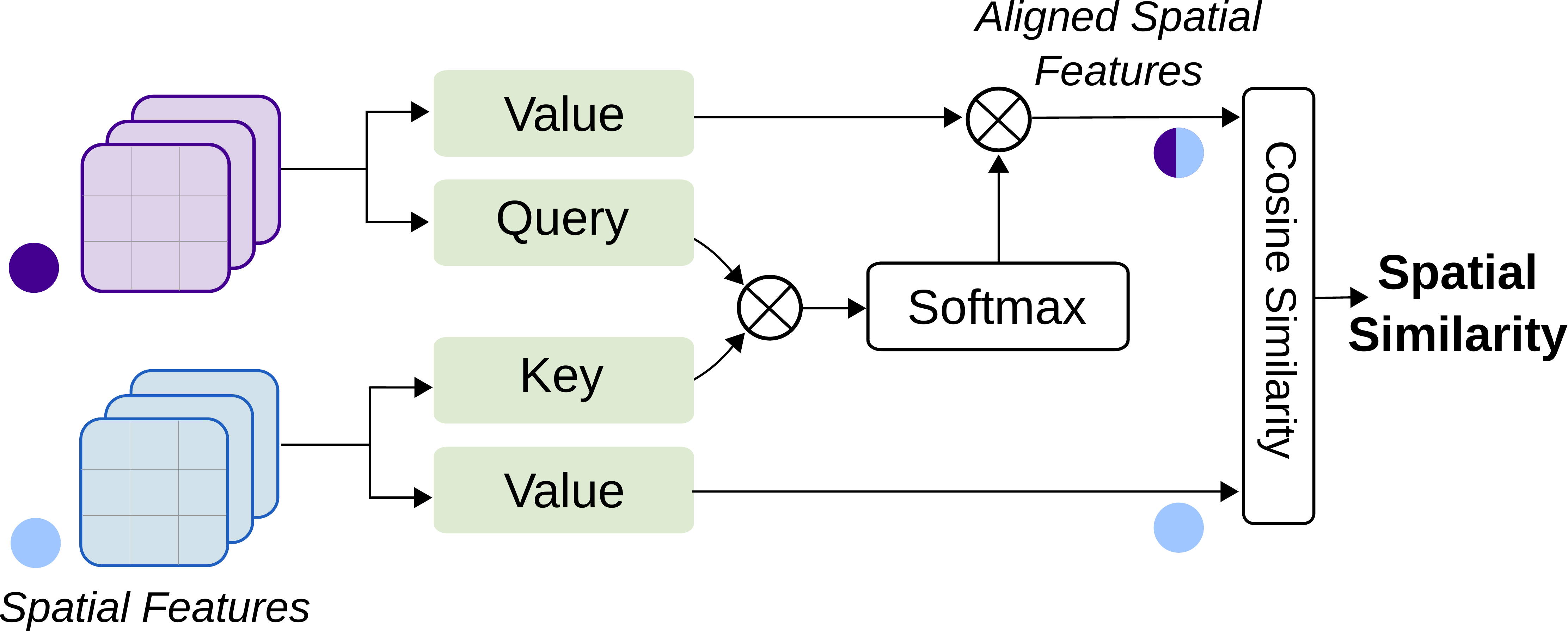}
  \vspace{-0.1in}
  \caption{{Attention-based Spatial Alignment.} To compute the spatial
  similarity between a pair of features ({\color{Purple}purple} and {\color{CornflowerBlue}blue}),
  we first spatially align the first features ({\color{Purple}purple}) with respect to 
  the second ({\color{CornflowerBlue}blue}) features with the attention mechanism (see \cref{eq:att}).
  Then we can compare the aligned value of the first features with
  the value of the second features.
  Note that the same process is applied in reverse to compute the final spatial similarity (see \cref{eq:sim}).}
  \label{fig:attention}
  \vspace{-0.15in}
\end{figure}

Although the GC loss is capable of producing good embeddings, using
the global features $\mbz^\rmg$ might suppress some local
discriminative features present in the spatial features $\mbz^\rms$ that can be informative
at the meta-testing stage (\eg suppressing object specific features
while overemphasizing the irrelevant background features).
Additionally, encoding the relevant spatial information into the learned representations
can play a critical role in
increasing the robustness of the embeddings and reducing their sensitivity to domain changes,
which is a highly desirable property for few-shot tasks.
To this end, we propose a novel SC loss as an alternative objective,
leveraging the spatial features $\mbz^\rms$
to compute the similarity between
a given pair of examples. However, to locally compare
a pair of spatial features $\mbz_i^\rms$ and $\mbz_j^\rms$ and compute the SC loss,
we first need to define a mechanism to align them spatially.
To this end, we employ the attention mechanism \cite{vaswani2017attention}
to compute the spatial attention weights to align the features $\mbz_i^\rms$
with respect to $\mbz_j^\rms$ and vice-versa. Then, we
measure the one-to-one spatial similarity as illustrated in \cref{fig:attention},
and finally, compute the SC loss.
a given pair of examples. However, to locally compare
a pair of spatial features $\mbz_i^\rms$ and $\mbz_j^\rms$ and compute the SC loss,
we first need to define a mechanism to align them spatially.
To this end, we employ the attention mechanism \cite{vaswani2017attention}
to compute the spatial attention weights to align the features $\mbz_i^\rms$
with respect to $\mbz_j^\rms$ and vice-versa. Then, we
measure the one-to-one spatial similarity as illustrated in \cref{fig:attention},
and finally, compute the SC loss.

\par \noindent \textbf{Attention-based Spatial Alignment.} 
Let $h_v$, $h_q$ and $h_k$ denote the value, query and key projection heads, taking as input the spatial features $\mbz^\rms$
and outputting the value $\mbv$, query $\mbq$ and key $\mbk$ of $d^{\prime}$-dimensional features,
\ie $\mbv, \mbq, \mbk \in \mathbb{R}^{HW \times d^{\prime}}$.
Given a pair of spatial features $\mbz_i^\rms$ and $\mbz_j^\rms$ of two
instances $i$ and $j$, we want to compute the aligned values of $i$ with respect to $j$, denoted as $\mbv_{i|j}$. Such an alignment can be 
obtained using the key $\mbk_i$ and the query $\mbq_j$ to compute the attention weights $\mba_{ij} \in \mathbb{R}^{HW \times HW}$, which can then
be applied to $\mbv_i$ to obtain $\mbv_{i|j}$. Concretely, this can be computed as follows:
\begin{equation}
\label{eq:att}
\mbv_{i|j} = \mba_{ij} \mbv_i \ \
\text{where}\ \ \mba_{ij} = \operatorname{softmax}\left(\frac{\mbq_j \mbk_i^{\top}}{\sqrt{d^{\prime}}}\right).
\end{equation}
Similarly, we compute $\mbv_{j|i}$ aligning the value of $j$
with respect to $i$ using the key $\mbk_j$ and the query $\mbq_i$.

\textit{Time Complexity.} The spatial alignment mechanism has a time complexity of $O(N^2 H^2 W^2 {d^{\prime}}^2)$, which
varies with the batch size, the size of the spatial features and the dimensionality of the values $\mbv$.
To avoid excessive cost, for large input images, we apply an adaptive average pooling
to reduce the size of the spatial features, in addition to using a small dimensionality $d^{\prime}$
and relatively small batches.

\par \noindent \textbf{Spatial Similarity.}
Given a pair of values $\mbv_i$ and $\mbv_j$, together with
their two aligned versions $\mbv_{i|j}$ and $\mbv_{j|i}$
computed using
the attention mechanism detailed above, and
with $\mbv_{*}^r$ denoting a feature vector at a spatial location $r \in [1,HW]$,
we first perform an $\ell_2$ normalization step of the values $\mbv_{*}^r$  at each spatial location $r$. Then,
we compute the total spatial similarity $\mathrm{sim}_{\mathrm{s}}(\mbz^\rms_i,\mbz^\rms_j)$
between a pair of spatial features
as follows:
\begin{equation}
\label{eq:sim}
\mathrm{sim}_{\mathrm{s}}(\mbz^\rms_i,\mbz^\rms_j)
= \frac{1}{HW} \sum_{r=1}^{HW} \left[ (\mbv^r_i)^{\top} \mbv^r_{j|i} + (\mbv^r_j)^{\top} \mbv^r_{i|j} \right].
\end{equation}

\par \noindent \textbf{Spatial Contrastive Learning.} With the spatial similarity function
$\mathrm{sim}_{\mathrm{s}}$ defined in \cref{eq:sim},
and similar to the GC loss in \cref{eq:GCL}, the SC loss can be computed as follows:
\begin{gather}
\label{eq:SCL}
L_{\rm{SC}} = \sum_{i=1}^{2N} \frac{1}{2N_{y_i} - 1} \sum_{j=1}^{2N} \mathbbm{1}_{i \neq j} \cdot \mathbbm{1}_{y_i=y_j} \cdot \ell_{ij}\ , \\
\text{where}\ \ \ell_{ij} = - \log \frac{\exp(\mathrm{sim}_{\mathrm{s}}(\mbz^\rms_i,\mbz^\rms_j) / \tau^\prime)}
{\sum_{k=1}^{2N} \mathbbm{1}_{i \neq k} \cdot \exp (\mathrm{sim}_{\mathrm{s}}(\mbz^\rms_i,\mbz^\rms_k) / \tau^\prime)}\ ,  \notag 
\end{gather}
with $\tau^\prime$ as a scalar temperature parameter.

\subsection{Pre-training Objective}
\label{subsec:total_loss}
Based on the contrastive objectives in \cref{eq:GCL} and \cref{eq:SCL}, the pre-training objective
can take different forms. We mainly consider the case where the pre-training objective $L_{\rm{T}}$ is the summation of
the CE and SC losses, with $\lambda_\mathrm{CE}$ and $\lambda_\mathrm{SC}$ as scaling weights to control
the contribution of each term:
\begin{equation}
\label{eq:total}
L_{\rmT} =  \lambda_\mathrm{CE} L_{\mathrm{CE}} + \lambda_\mathrm{SC} L_{\mathrm{SC}}.
\end{equation}
However, we also explore other alternatives such as replacing $L_\mathrm{SC}$ with $L_\mathrm{GC}$ or training with
both $L_\mathrm{GC}$ and $L_\mathrm{SC}$ as auxiliary losses with their corresponding weighting terms.
Additionally, we also consider the self-supervised formulations of the GC and SC losses, where
the label information is discarded and the only positives considered are the augmented versions of each
example (\ie $y_i = i\mod{N}$). We refer to them as SS-GC and SS-SC
(Self-Supervised Global and Spatial Contrastive) losses respectively.

Using the total loss $L_{\rmT}$, the embedding model $f_\phi$ can be trained together with
the projection head and the attention modules during the meta-training stage.
Specifically, let $\psi$ represent the parameters of the projection head $p$ and the attention modules $h_v$, $h_q$ and $h_k$.
The parameters are obtained as follows:
\begin{equation}
\{\phi, \psi\} = \underset{\{\phi, \psi\}}{\arg\min}\ L_{\mathrm{T}} (\cD^\mathrm{new} ; \{\phi, \psi\}).
\end{equation}
After the pre-training stage, the parameters $\psi$ are discarded, and the embedding model $f_\phi$ is then fixed
and carried over from meta-training to meta-testing.

\begin{figure}[t]
  \centering
  \includegraphics[width=0.65\linewidth]{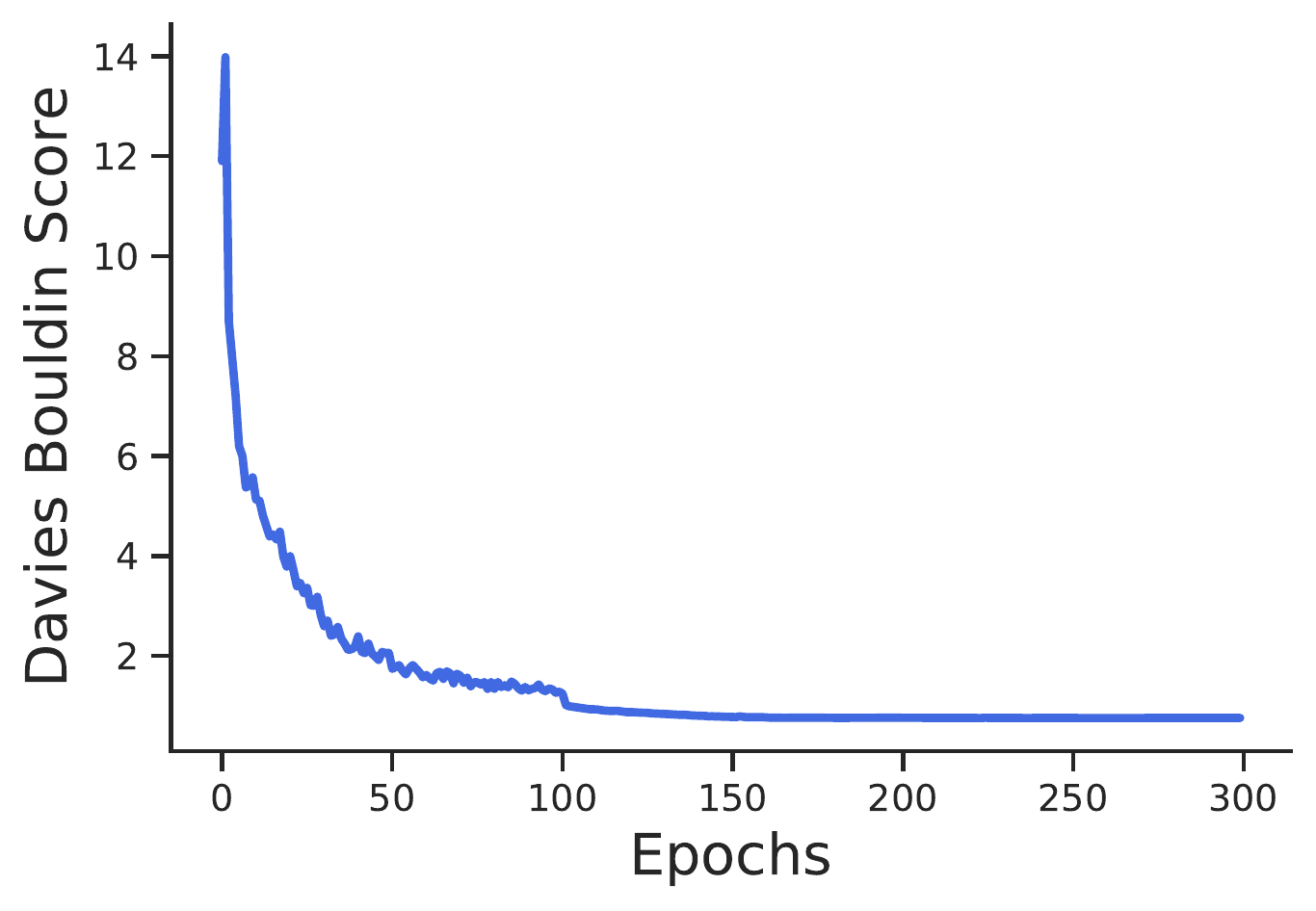}
  \vspace{-0.1in}
  \caption{{Degree of Clustering.} The plot shows the evolution of the intra-class variation using the
  Davies-Bouldin index \cite{davies1979cluster} during the course of training on \textit{mini}-ImageNet when using the contrastive
  loss. We see that the learned embeddings of each class are significantly over-clustered, an outcome that might not
  be desired in some cases.}
  \label{fig:oveclustering}
  \vspace{-0.2in}
\end{figure}

\subsection{Avoiding Excessive Disentanglement}
\label{subsec:regulatization}

Since the contrastive objectives encourage closely aligned
embeddings of instances of the same class while
distributing all of the normalized features uniformly on the hypersphere \cite{wang2020understanding},
we have to consider a possible over-clustering of the features of the same class (see \cref{fig:oveclustering}).
Such an outcome can be desired for closed-set recognition, but in a few-shot setting, in which the discrepancy
between the meta-training and meta-testing domain might differ greatly from one case to the other
(\eg training on coarse seen categories, and testing
on fine-grained unseen sub-categories), this might lead to sub-optimal performances.
As such, to avoid an excessive disentanglement
of the learned features and to further improve the generalization of the embedding model,
we propose Contrastive Distillation (CD) to reduce the compactness
of the features in embeddings space.

\par \noindent \textbf{Contrastive Distillation.} 
Given a teacher model $f_{\phi_t}$
pre-trained with the objective in \cref{eq:total},
we transfer its knowledge to a student model $f_{\phi_s}$ using
the standard knowledge distillation \cite{hinton2015distilling} objective $L_{\mathrm{KL}}$
(\ie the Kullback-Leibler (KL) divergence between the student's predictions and
the soft targets predicted by the teacher), but with an additional
contrastive distillation loss $L_\mathrm{CD}$.
This loss consists of maximizing the inner dot product between the $\ell_2$ normalized global
features of the teacher $\mbz^{\rmg t}$ and that of the student $\mbz^{\rmg s}$, which
corresponds to minimizing the squared Euclidean distance, formally:
\begin{equation}
L_{\mathrm{CD}} = \frac{1}{N} \sum_{i=1}^{N} \|\mbz^{\rmg t}_i - \mbz^{\rmg s}_i \|_{2}^{2}.
\end{equation}
To summarize, the student's parameters are learned as follows:
\begin{equation}
\begin{split}
\phi_s = \underset{\phi_s}{\arg\min}\ &\lambda_\mathrm{CD} L_{\mathrm{CD}}(\cD^\mathrm{new} ; \phi_s, \phi_t)\ + \\ &\lambda_\mathrm{KL} L_{\mathrm{KL}}(\cD^\mathrm{new} ; \phi_s, \phi_t).
\end{split}
\end{equation}
As result, and different from the standard contrastive distillation loss
\cite{tian2019contrastivedistill}
that leverages negative samples, only maximizing the similarity between the pairs of features
without using any negatives relaxes the uniformity constraint of
the contrastive loss, which in turns reduces the disentanglement of the learned embeddings.

\section{Experiments}
\label{sec:exp}
For the experimental section, we base our implementation on the publicly available code of \cite{tian2020rethinking} and
conduct experiments on four popular few-shot classification benchmarks:
\textit{mini}-ImageNet \cite{vinyals2016matching}, \textit{tiered}-ImageNet \cite{ren2018meta}, CIFAR-CS \cite{bertinetto2018meta}
and FC100 \cite{oreshkin2018tadam}. Additionally, we present experiments on
cross-domain few-shot benchmarks introduced by \cite{tseng2020cross}. 
We note that additional experimental details and results are
presented in the supplementary material.

\subsection{Experimental Details}

\par \noindent \textbf{Architecture.} For the embedding model $f_\phi$, we follow \cite{tian2020rethinking}
and use a ResNet-12 consisting of 4 residual blocks with
Dropblock as a regularizer and 640-dimensional output features (\ie $d=$ 640).
For the projection head and the attention modules, we use an MLP with one
hidden layer and a ReLU non-linearity similar to SimCLR,
outputting 80-dimensional features (\ie $d^\prime=$ 80).

\par \noindent \textbf{Training Setup.} For optimization, we use SGD with
a momentum of 0.9, a weight decay of $5 \times 10^{-4}$, a learning rate of $5 \times 10^{-2}$ and a batch size of $64$.
For the loss functions, we set the temperature parameters $\tau$ and $\tau^\prime$ to $0.1$
and the scaling weights $\lambda_\mathrm{CE}$, $\lambda_\mathrm{SC}$, and $\lambda_\mathrm{GC}$ to 1.0,
except for CIFAR-FS where we set them to 0.5. For distillation,
we set $\lambda_\mathrm{CD}$ to 10.0 and $\lambda_\mathrm{KL}$ to 1.0 and use a temperature of 4.0
for the KL loss. %

\par \noindent \textbf{Data Augmentation.}
During meta-training, for a given augmented batch of $2N$ examples, and consistent with
other approaches \cite{tian2020rethinking,lee2019meta}, the first $N$  instances
are obtained using standard augmentations, \ie random crop, color jittering and random horizontal flip.
The remaining $N$ instances are obtained with SimCLR type augmentations, \ie random resized crop,
color jittering, random horizontal flip and
random grayscale conversion. During the meta-testing stage, we follow \cite{tian2020rethinking} and create 5 augmented versions
of each training image to overcome the problem of data insufficiency and train the linear classifier $f_{\theta}$.

\par \noindent \textbf{Evaluation Setup.}
During meta-testing, and given a pre-trained embedding model $f_\phi$, we follow \cite{tian2020rethinking}
and consider a linear classifier as the predictor
$f_{\theta}$, implemented in  scikit-learn \cite{pedregosa2011scikit}
and trained on the $\ell_2$ normalized features produced by $f_\phi$.
Specifically, we sample a number of $C$-way $K$-shot testing classification tasks
constructed from the unseen classes of the meta-testing set, with $C$ as the number of classes
and $K$ as the number of training examples per class.
After training $f_{\theta}$ on the train set, the predictor is then applied to
the features of the test set to obtain the prediction and compute the accuracy.
In our case, we evaluate the model over 600 randomly sampled tasks and report
the median accuracy over 3 runs with 95\% confidence intervals, where in each
run, the accuracy is the mean accuracy of the 600 sampled tasks.

\subsection{Ablation Studies}

\begin{table}[!t]
\centering
\small
  \scalebox{0.72}{
  \begin{tabular}{lccccc}
  \toprule
  \multirow{2}{*}{Loss Function} & \multirow{2}{*}{Aug.} & \multicolumn{2}{c}{\textit{mini}-ImageNet, 5-way} & \multicolumn{2}{c}{CIFAR-CS, 5-way} \\
                     &                       & 1-shot & 5-shot                   & 1-shot & 5-shot                \\
  \midrule
  CE            &                                     & 61.8 $\pm$ 0.7 &  79.7 $\pm$ 0.6 & 71.3 $\pm$ 0.9 & 86.1 $\pm$ 0.6 \\
  CE            & $\checkmark$                        & 61.8 $\pm$ 0.8 &  78.6 $\pm$ 0.5 & 71.9 $\pm$ 0.9 & 86.3 $\pm$ 0.5 \\
  CE + SS-GC  & $\checkmark$                          & 62.7 $\pm$ 0.7 &  81.0 $\pm$ 0.6 & 70.9 $\pm$ 0.9 & 84.5 $\pm$ 0.6 \\
  CE + SS-SC  & $\checkmark$                          & 64.0 $\pm$ 0.8 &  81.5 $\pm$ 0.5 & 72.1 $\pm$ 0.8 & 86.2 $\pm$ 0.6 \\
  CE + SS-GC + SS-SC  & $\checkmark$                  & 62.8 $\pm$ 0.8 &  81.1 $\pm$ 0.6 & 69.0 $\pm$ 0.9 & 85.0 $\pm$ 0.6 \\
  CE + GC    & $\checkmark$                           & 65.0 $\pm$ 0.8 &  81.6 $\pm$ 0.5 & 74.0 $\pm$ 0.8 & 87.3 $\pm$ 0.6 \\
  CE + SC    & $\checkmark$                           & \bt{65.7 $\pm$ 0.8} &  \bt{82.5 $\pm$ 0.5} & 75.0 $\pm$ 0.9 & 87.4 $\pm$ 0.6 \\
  CE + GC + SC  & $\checkmark$                        & 65.0 $\pm$ 0.8 &  81.3 $\pm$ 0.5 & \bt{76.0 $\pm$ 0.7} & \bt{87.5 $\pm$ 0.5} \\
  \bottomrule
  \end{tabular}}
  \vspace{-0.1in}
  \caption{Comparison of the mean acc. obtained on \textit{mini}-ImageNet and CIFAR-FS with different training objectives.
  ``Aug.'' indicates the usage of SimCLR type augmentations.}%
  \label{tab:ablationLoss}
\end{table}

\begin{figure}[t]
  \centering
  \includegraphics[width=0.9\linewidth]{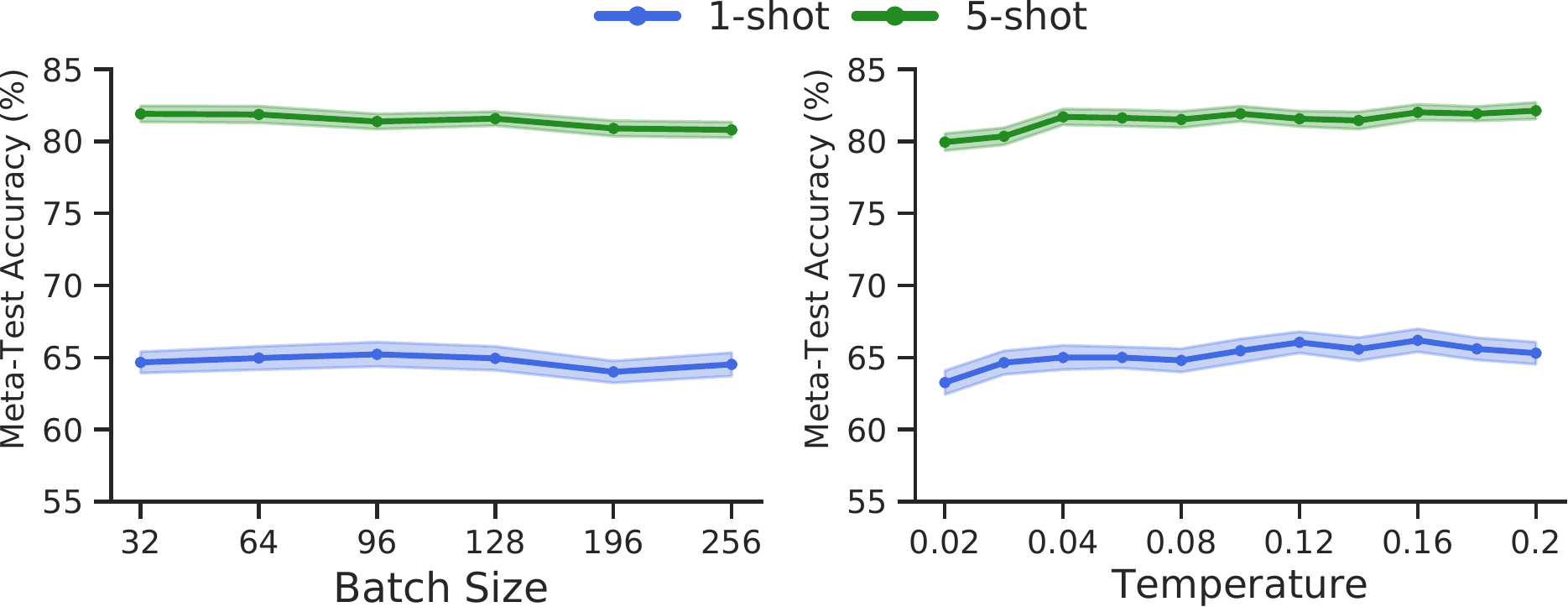}
  \vspace{-0.1in}
  \caption{Comparison of the mean acc. obtained on \textit{mini}-ImageNet across various batch sizes
  and SC loss temperatures.}
  \label{fig:hyperparameters}
\end{figure}

\begin{table}[!t]
\centering
\small
  \scalebox{0.83}{
  \begin{tabular}{lcccc}
  \toprule
  \multirow{2}{*}{Augmentation} & \multicolumn{2}{c}{\textit{mini}-ImageNet, 5-way} & \multicolumn{2}{c}{CIFAR-CS, 5-way} \\
                               & 1-shot & 5-shot                   & 1-shot & 5-shot                \\
  \midrule
  Standard                     & 64.3 $\pm$ 0.7 &  80.6 $\pm$ 0.5 & 74.9 $\pm$ 0.8 & 86.3 $\pm$ 0.6 \\
  SimCLR                       & \bt{65.7 $\pm$ 0.8} &  \bt{82.5 $\pm$ 0.5} & \bt{75.0 $\pm$ 0.9} & 87.4 $\pm$ 0.6 \\
  AutoAugment                  & 65.2 $\pm$ 0.7 &  82.1 $\pm$ 0.5 & 74.0 $\pm$ 0.9 & 86.7 $\pm$ 0.6 \\
  Stacked RandAug.             & 64.9 $\pm$ 0.8 &  81.6 $\pm$ 0.6 & \bt{75.0 $\pm$ 0.9} & \bt{87.6 $\pm$ 0.6} \\
  \bottomrule
  \end{tabular}}
  \vspace{-0.1in}
  \caption{Comparison of the mean acc. obtained on \textit{mini}-ImageNet and CIFAR-FS with different
  augmentation strategies, which are used to obtain the additional $N$ augmented instances within
  a minibatch.}
  \label{tab:ablationAug}
\end{table}

\begin{table}[!t]
\centering
\small
  \scalebox{0.85}{
  \begin{tabular}{lccccc}
  \toprule
  \multirow{2}{*}{Aggregation} && \multicolumn{2}{c}{\textit{mini}-ImageNet, 5-way} & \multicolumn{2}{c}{CIFAR-CS, 5-way} \\
                   && 1-shot & 5-shot                   & 1-shot & 5-shot              \\
  \midrule
  Sum                  && 65.2 $\pm$ 0.8 &  81.2 $\pm$ 0.5 & \bt{75.3 $\pm$ 0.8} & 87.3 $\pm$ 0.5 \\
  Mean                 && \bt{65.7 $\pm$ 0.8} &  \bt{82.5 $\pm$ 0.5} & 75.0 $\pm$ 0.9 & \bt{87.4 $\pm$ 0.6} \\
  Maximum              && 65.5 $\pm$ 0.7 &  82.0 $\pm$ 0.5 & 73.4 $\pm$ 0.8 & 86.4 $\pm$ 0.6 \\
  LogSumExp            && 64.8 $\pm$ 0.8 &  81.7 $\pm$ 0.6 & 74.2 $\pm$ 0.8 & 87.0 $\pm$ 0.6 \\
  \bottomrule
  \end{tabular}}
  \vspace{-0.10in}
  \caption{
  Comparison of the mean acc. obtained on \textit{mini}-ImageNet and CIFAR-FS with different
  aggregation functions, which are used to amount the total similarity
  from the one-to-one spatial similarities.}
  \label{tab:ablationAgg}
\end{table}

\begin{table}[!t]
\centering
\small
  \scalebox{0.8}{
  \begin{tabular}{lcccc}
  \toprule
  \multirow{2}{*}{Features Used} & \multicolumn{2}{c}{\textit{mini}-ImageNet, 5-way} & \multicolumn{2}{c}{CIFAR-CS, 5-way} \\
                               & 1-shot & 5-shot                   & 1-shot & 5-shot                \\
  \midrule
  Spatial                    & 64.5 $\pm$ 0.8 &  82.1 $\pm$ 0.5 & \bt{75.0 $\pm$ 0.9} & 87.1 $\pm$ 0.6 \\
  Global                     & 65.7 $\pm$ 0.8 &  82.5 $\pm$ 0.5 & \bt{75.0 $\pm$ 0.9} & 87.4 $\pm$ 0.6 \\
  Glo. \& Spa. (Max)         & 65.6 $\pm$ 0.8 &  82.1 $\pm$ 0.5 & 74.2 $\pm$ 0.8 & 87.3 $\pm$ 0.5 \\
  Glo. \& Spa. (Sum)         & \bt{65.7 $\pm$ 0.8} &  \bt{83.1 $\pm$ 0.5} & \bt{75.6 $\pm$ 0.9} & \bt{87.6 $\pm$ 0.6} \\
  \bottomrule
  \end{tabular}}
  \vspace{-0.1in}
  \caption{Comparison of the mean acc. obtained on \textit{mini}-ImageNet and CIFAR-FS with
  different evaluation settings, in which we use either the global features, the spatial features,
  or both.}
  \label{tab:ablationEval}
\end{table}

\begin{table}[!t]
\centering
\small
  \scalebox{0.82}{
  \begin{tabular}{lcccc}
  \toprule
  \multirow{2}{*}{Loss Function} & \multicolumn{2}{c}{\textit{mini}-ImageNet, 5-way} & \multicolumn{2}{c}{CIFAR-CS, 5-way} \\
                               & 1-shot & 5-shot                   & 1-shot & 5-shot                \\
  \midrule
  \textit{Teacher}             & \textit{65.7 $\pm$ 0.8} &  \textit{82.5 $\pm$ 0.5} & \textit{75.0 $\pm$ 0.9} & \textit{87.4 $\pm$ 0.6} \\
  \addlinespace[0.05in]
  KL                           & 66.0 $\pm$ 0.8 & 82.5 $\pm$ 0.5 & 75.9 $\pm$ 0.9 & 87.4 $\pm$ 0.6 \\
  KL+CD                        & \textbf{67.4 $\pm$ 0.8} & \textbf{82.7 $\pm$ 0.5} & \textbf{76.5 $\pm$ 0.9} & \textbf{87.6 $\pm$ 0.6} \\
  \bottomrule
  \end{tabular}}
  \vspace{-0.1in}
  \caption{Comparison of the mean acc. obtained on \textit{mini}-ImageNet and CIFAR-FS with different
  distillation objectives.}
  \label{tab:ablationDistill}
\end{table}

\begin{figure}[t]
  \centering
  \includegraphics[width=0.9\linewidth]{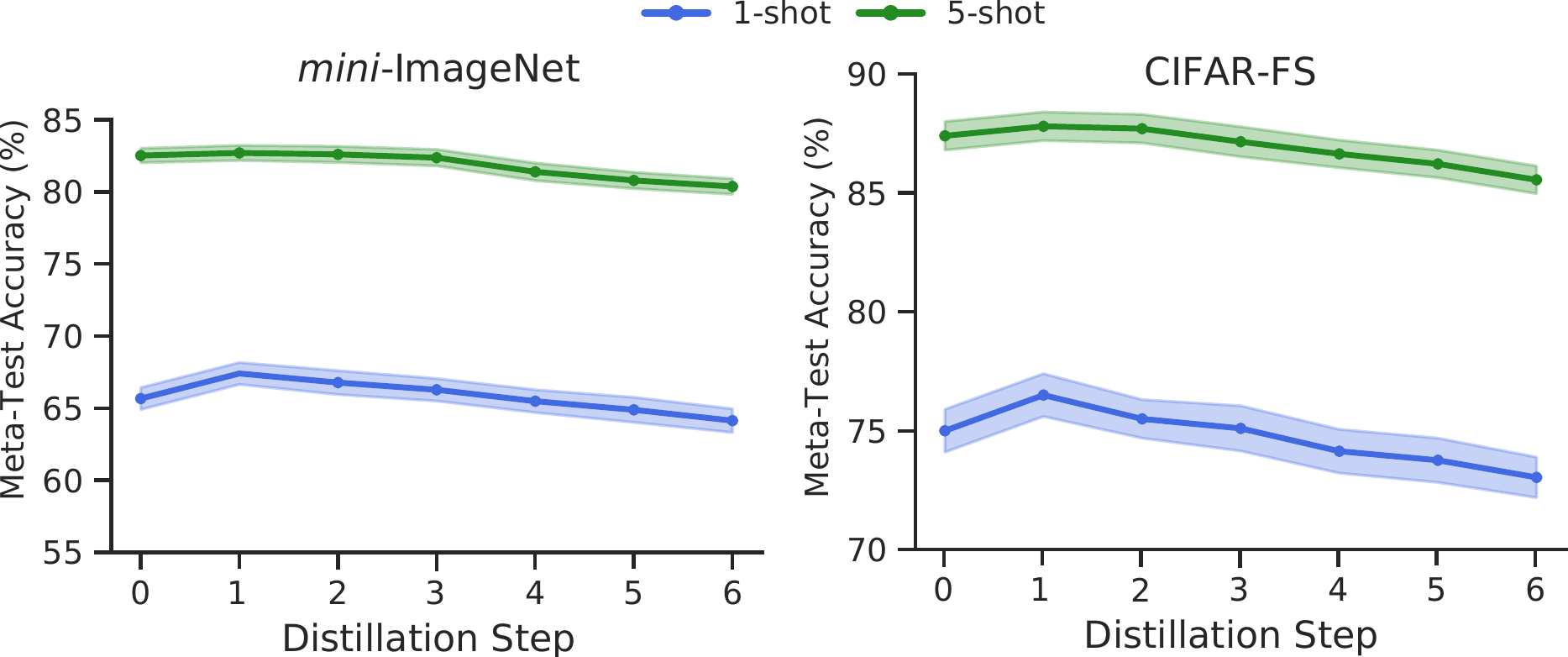}
  \vspace{-0.1in}
  \caption{Comparison of the mean acc. obtained on \textit{mini}-ImageNet and CIFAR-FS with sequential distillation.}
  \label{fig:ablationDistill}
  \vspace{-0.2in}
\end{figure}

\begin{table*}[!t]
\setlength{\tabcolsep}{0.3cm}
	\small
	\centering
	\scalebox{0.9}{
	\begin{tabular}{l l c c c c}
		\toprule
		& & \multicolumn{2}{c}{\textit{mini}-ImageNet, 5-way} & \multicolumn{2}{c}{\textit{tiered}-ImageNet, 5-way} \\
		\multirow{-2}{*}{Method} & \multirow{-2}{*}{Backbone} & 1-shot & 5-shot &  1-shot & 5-shot  \\ \midrule
		MAML~\cite{finn2017model}                      & 32-32-32-32   & 48.70 $\pm$ 1.84    & 63.11 $\pm$ 0.92    &  51.67 $\pm$ 1.81  &  70.30 $\pm$ 1.75 \\
		Matching Networks~\cite{vinyals2016matching}   & 64-64-64-64   & 43.56 $\pm$ 0.84    & 55.31 $\pm$ 0.73    & \ \ - & \ \ - \\
		Prototypical Networks$^\dagger$~\cite{snell2017prototypical}   & 64-64-64-64  & 49.42 $\pm$ 0.78 & 68.20 $\pm$ 0.66 & 53.31 $\pm$ 0.89 & 72.69 $\pm$ 0.74 \\
		Relation Networks~\cite{sung2018learning}      & 64-96-128-256 & 50.44 $\pm$ 0.82    & 65.32 $\pm$ 0.70    & 54.48 $\pm$ 0.93 & 71.32 $\pm$ 0.78 \\
		SNAIL~\cite{mishra2018a}                       & ResNet-12     & 55.71 $\pm$ 0.99    & 68.88 $\pm$ 0.92    & \ \ - & \ \ - \\
		TADAM~\cite{oreshkin2018tadam}                 & ResNet-12     & 58.50 $\pm$ 0.30    & 76.70 $\pm$ 0.30    & \ \ - & \ \ - \\
		Shot-Free~\cite{ravichandran2019few}           & ResNet-12     & 59.04 $\pm$ n/a \ \ & 77.64 $\pm$ n/a \ \ & 63.52 $\pm$ n/a \ \ & 82.59 $\pm$ n/a \ \ \\
		MetaOptNet~\cite{lee2019meta}                  & ResNet-12     & 62.64 $\pm$ 0.61    & 78.63 $\pm$ 0.46    & 65.99 $\pm$ 0.72 & 81.56 $\pm$ 0.53 \\
		Diversity w/ Coop.~\cite{dvornik2019diversity} & ResNet-18& 59.48 $\pm$ 0.65    & 75.62 $\pm$ 0.48    & \ \ - & \ \ - \\
		Boosting~\cite{gidaris2019boosting}            & WRN-28-10     & 63.77 $\pm$ 0.45    & 80.70 $\pm$ 0.33    & 70.53 $\pm$ 0.51 & 84.98 $\pm$ 0.36  \\
		Fine-tuning~\cite{dhillon2019baseline}         & WRN-28-10     & 57.73 $\pm$ 0.62    & 78.17 $\pm$ 0.49    & 66.58 $\pm$ 0.70 & 85.55 $\pm$ 0.48 \\
		LEO-trainval$^\dagger$~\cite{rusu2018meta} & WRN-28-10 & 61.76 $\pm$ 0.08    & 77.59 $\pm$ 0.12    & 66.33 $\pm$ 0.05 & 81.44 $\pm$ 0.09 \\
		RFS~\cite{tian2020rethinking}              & ResNet-12     & 62.02 $\pm$ 0.63    & 79.64 $\pm$ 0.44    & 69.74 $\pm$ 0.72 & 84.41 $\pm$ 0.55 \\
		RFS-Distill~\cite{tian2020rethinking}      & ResNet-12     & 64.82 $\pm$ 0.60    & 82.14 $\pm$ 0.43    & 71.52 $\pm$ 0.69 & 86.03 $\pm$ 0.49 \\
    \midrule
    \rowcolor{lightgray} Ours                 & ResNet-12     & 65.69 $\pm$ 0.81          & 83.10 $\pm$ 0.52           & 71.48 $\pm$ 0.89   & \textbf{86.88 $\pm$ 0.53} \\
    \rowcolor{lightgray} Ours-Distill         & ResNet-12     & \textbf{67.40 $\pm$ 0.76} & \textbf{83.19 $\pm$ 0.54} & \textbf{71.98 $\pm$ 0.91}   & 86.19 $\pm$ 0.59 \\
		\bottomrule
	\end{tabular}}
	\vspace{-0.1in}
	\caption{Comparison with prior few-shot classification works on ImageNet derivatives.
	We show the mean acc. and 95\% confidence interval.
	$^\dagger$results obtained by training on both train and validation sets.}
	\label{tab:resultsImageNet}
\end{table*}

\begin{table*}[!t]
\setlength{\tabcolsep}{0.4cm}
	\small
	\centering
	\scalebox{0.9}{
		\begin{tabular}{l l c c c c}
		\toprule
		&  & \multicolumn{2}{c}{CIFAR-FS, 5-way} & \multicolumn{2}{c}{FC100, 5-way} \\
		\multirow{-2}{*}{Method} & \multirow{-2}{*}{Backbone} & 1-shot & 5-shot &  1-shot & 5-shot  \\ \midrule
		MAML~\cite{finn2017model}                      & 32-32-32-32   & 58.9 $\pm$ 1.9    & 71.5 $\pm$ 1.0 &  \ \ - & \ \ - \\
		Relation Networks~\cite{sung2018learning}      & 64-96-128-256 & 55.0 $\pm$ 1.0    & 69.3 $\pm$ 0.8 & \ \ - & \ \ - \\
		R2D2~\cite{bertinetto2018meta}                 & 96-192-384-512& 65.3 $\pm$ 0.2    & 79.4 $\pm$ 0.1 & \ \ - & \ \ - \\
		TADAM~\cite{oreshkin2018tadam}                 & ResNet-12     & \ \ -             & \ \ -          & 40.1 $\pm$ 0.4 & 56.1 $\pm$ 0.4 \\
		Shot-Free~\cite{ravichandran2019few}           & ResNet-12     & 69.2 $\pm$ n/a    & 84.7 $\pm$ n/a & \ \ - & \ \ - \\
		TEWAM~\cite{qiao2019transductive}              & ResNet-12     & 70.4 $\pm$ n/a    & 81.3 $\pm$ n/a & \ \ - & \ \ - \\
		Prototypical Networks$^\dagger$~\cite{snell2017prototypical}   & ResNet-12  & 72.2 $\pm$ 0.7 & 83.5 $\pm$ 0.5 & 37.5 $\pm$ 0.6 & 52.5 $\pm$ 0.6 \\
		Boosting~\cite{gidaris2019boosting}            & WRN-28-10     & 73.6 $\pm$ 0.3    & 86.0 $\pm$ 0.2 & \ \ - & \ \ - \\
		MetaOptNet~\cite{lee2019meta}                  & ResNet-12     & 72.6 $\pm$ 0.7    & 84.3 $\pm$ 0.5 & 41.1 $\pm$ 0.6 & 55.5 $\pm$ 0.6 \\
		RFS~\cite{tian2020rethinking}                  & ResNet-12     & 71.5 $\pm$ 0.8    & 86.0 $\pm$ 0.5 & 42.6 $\pm$ 0.7 & 59.1 $\pm$ 0.6 \\
		RFS-Distill~\cite{tian2020rethinking}          & ResNet-12     & 73.9 $\pm$ 0.8    & 86.9 $\pm$ 0.5 & 44.6 $\pm$ 0.7 & 60.9 $\pm$ 0.6 \\
    \midrule
    \rowcolor{lightgray} Ours                 & ResNet-12     &  75.6 $\pm$ 0.9    & 87.6 $\pm$ 0.6    & 44.4 $\pm$ 0.8   & 60.8 $\pm$ 0.8 \\
    \rowcolor{lightgray} Ours-Distill         & ResNet-12     &  \textbf{76.5 $\pm$ 0.9}    & \textbf{88.0 $\pm$ 0.6}    & \textbf{44.8 $\pm$ 0.7}   & \textbf{61.4 $\pm$ 0.7} \\
    \bottomrule
	\end{tabular}}
	\vspace{-0.1in}
	\caption{Comparison with prior few-shot classification works on CIFAR-10 derivatives.
	We show the mean acc. and 95\% confidence interval.
	$^\dagger$results obtained by training on both train and validation sets.}
	\label{tab:resultsCIFAR}
	\vspace{-0.1in}
\end{table*}

\begin{table}[!t]
  \small
  \centering
  \scalebox{0.85}{
  \begin{tabular}{lc}
    \toprule
    \multirow{1}{*}{Method} & \textit{mini}-ImageNet   \\
                        &  20-way 1-shot  \\
    \midrule
    Matching networks~\cite{vinyals2016matching}     & 17.31 $\pm$ 0.22  \\
    Meta-LSTM~\cite{ravi2016optimization}            & 16.70 $\pm$ 0.23 \\
    MAML~\cite{finn2017model}                        & 16.49 $\pm$ 0.58 \\
    Meta-SGD~\cite{li2017meta}                       & 17.56 $\pm$ 0.64 \\
    LGM-Net~\cite{li2019lgm}                         & 26.14 $\pm$ 0.34 \\    
    \midrule
    \rowcolor{lightgray} Ours                        & 36.81 $\pm$ 0.38 \\
    \rowcolor{lightgray} Ours-Distill                & \textbf{37.47 $\pm$ 0.32} \\
    \bottomrule
  \end{tabular}}
	\vspace{-0.1in}
	\caption{Comparison with prior few-shot classification works on 20-way 1-shot \textit{mini}ImageNet
	classification. We show the mean acc. and 95\% confidence
	interval.}
	\label{tab:resultsMini20way}
	\vspace{-0.2in}
\end{table}

\begin{table*}[!t]
  \small
  \centering
    \scalebox{0.815}{
    \begin{tabular}{lcccccccc}
    \toprule
    \multirow{2}{*}{Method} & \multicolumn{2}{c}{CUB, 5-way} & \multicolumn{2}{c}{Cars, 5-way}  & \multicolumn{2}{c}{Places, 5-way} & \multicolumn{2}{c}{Plantae, 5-way} \\
                             & 1-shot & 5-shot & 1-shot & 5-shot & 1-shot & 5-shot & 1-shot & 5-shot \\
    \midrule
    MatchingNet~\cite{vinyals2016matching}          & 35.89 $\pm$ 0.5 & 51.37 $\pm$ 0.7 & 30.77 $\pm$ 0.5 & 38.99 $\pm$ 0.6 & 49.86 $\pm$ 0.8 & 63.16 $\pm$ 0.8 & 32.70 $\pm$ 0.6 & 46.53 $\pm$ 0.6 \\
    MatchingNet w/ FT~\cite{tseng2020cross}        & 36.61 $\pm$ 0.6 & 55.23 $\pm$ 0.8 & 29.82 $\pm$ 0.4 & 41.24 $\pm$ 0.6 & 51.07 $\pm$ 0.7 & 64.55 $\pm$ 0.7 & 34.48 $\pm$ 0.5 & 41.69 $\pm$ 0.6 \\
    RelationNet~\cite{sung2018learning}       & 42.44 $\pm$ 0.7 & 57.77 $\pm$ 0.7 & 29.11 $\pm$ 0.6 & 37.33 $\pm$ 0.7 & 48.64 $\pm$ 0.8 & 63.32 $\pm$ 0.8 & 33.17 $\pm$ 0.6 & 44.00 $\pm$ 0.6 \\
    RelationNet w/ FT~\cite{tseng2020cross}        & 44.07 $\pm$ 0.7 & 59.46 $\pm$ 0.7 & 28.63 $\pm$ 0.6 & 39.91 $\pm$ 0.7 & 50.68 $\pm$ 0.9 & 66.28 $\pm$ 0.7 & 33.14 $\pm$ 0.6 & 45.08 $\pm$ 0.6 \\
    GNN~\cite{garcia2017few}                        & 45.69 $\pm$ 0.7 & 62.25 $\pm$ 0.6 & 31.79 $\pm$ 0.5 & 44.28 $\pm$ 0.6 & 53.10 $\pm$ 0.8 & 70.84 $\pm$ 0.6 & 35.60 $\pm$ 0.5 & 52.53 $\pm$ 0.6 \\
    GNN w/ FT~\cite{tseng2020cross}                & 47.47 $\pm$ 0.6 & 66.98 $\pm$ 0.7 & 31.61 $\pm$ 0.5 & 44.90 $\pm$ 0.6 & 55.77 $\pm$ 0.8 & 73.94 $\pm$ 0.7 & 35.95 $\pm$ 0.5 & 53.85 $\pm$ 0.6 \\
    \midrule
    \rowcolor{lightgray} Ours                 & 49.58 $\pm$ 0.7 & 67.64 $\pm$ 0.7 & 34.46 $\pm$ 0.6 & \bt{52.22 $\pm$ 0.7} & 59.37 $\pm$ 0.7 & 76.46 $\pm$ 0.6 & \bt{40.23 $\pm$ 0.6} & 59.38 $\pm$ 0.6 \\
    \rowcolor{lightgray} Ours-Distill         & \bt{50.09 $\pm$ 0.7} & \bt{68.81 $\pm$ 0.6} & \bt{34.93 $\pm$ 0.6} & 51.72 $\pm$ 0.7 & \bt{60.32 $\pm$ 0.8} & \bt{76.51 $\pm$ 0.6} & 39.75 $\pm$ 0.8 & \bt{59.91 $\pm$ 0.6} \\
    \bottomrule
  \end{tabular}}
  \vspace{-0.1in}
  \caption{Comparison with prior works on cross-domain few-shot classification benchmarks.
  We train the model on the \textit{mini}-ImageNet domain and evaluate the trained model on
  other domains. We show the mean acc. and 95\% confidence interval.}
  \label{tab:resultsCrossDomain}
  \vspace{-0.1in}
\end{table*}

We start by conducting detailed ablation studies to analyze the contribution of each
component of the proposed method, from the choices of the loss function to the hyperparameters of the SC loss.

\par \noindent \textbf{Loss Functions.}
To investigate the effect of the contrastive losses when used as auxiliary training objectives, we evaluate the performances
obtained with various loss functions as detailed in \cref{subsec:total_loss}.
The results are shown in \cref{tab:ablationLoss}. We observe a notable gain in performance when adopting auxiliary contrastive losses,
be it supervised or self-supervised, with better gains when using the supervised formulation, highlighting the benefits of
using the label information when constructing the positives and negatives samples. More importantly, the SC loss
outperforms the standard GC loss, confirming the effectiveness of using the spatial features
rather than the global features. Additionally, using both the SC and GC losses does not result in distinct gains over the
SC loss. Thus, for the rest of this section, we adopt the SC as a sole auxiliary loss.

\par \noindent \textbf{Spatial Contrastive Loss.}
In this section, we examine different variations and hyperparameters of the SC loss when used as
an auxiliary objective along the CE loss. In particular, we consider the following variations:

\par \noindent \textit{- Hyperparameters.}
To inspect the SC loss's hyperparameter stability, %
we conduct experiments with different batch sizes and temperature values. As seen in \cref{fig:hyperparameters}, by combining
the CE and the SC losses, we leverage the stability of the CE and obtain consistent results across several batch sizes,
circumventing the need for very large batches
when training with only the contrastive losses as it is the case in the unsupervised representation learning setting.
As for the temperatures, disregarding the low temperatures in which the SC loss is dominated
by the small distances, rendering the actual distances between widely separated representations
almost irrelevant, we see comparable performances for temperatures above 0.05, further confirming the stability of the approach.

\par \noindent \textit{- Augmentations.} Although we mainly use SimCLR type augmentations to produce the additional $N$ augmented examples
within a given batch, other augmentations can also be used. Specifically, we consider the standard augmentations
used when training with only the CE loss, AutoAugment \cite{Cubuk2019CVPR} and Stacked RandAugment \cite{tian2020makes}.
\cref{tab:ablationAug} shows that the SimCLR type augmentations yield the best results overall.
We speculate that for the standard augmentation, without any novel transformations that the model is forced to be invariant under, the gains
are minimal. As for strong augmentations (\ie AutoAugment and Stacked RandAugment), the augmented inputs might be substantially deformed, making
the spatial alignment insufficient and reducing the effect of the SC loss.

\par \noindent \textit{- Aggregation Function.}
\cref{tab:ablationAgg} presents the results obtained with various aggregation functions used to aggregate the one-to-one spatial similarities
into an overall measure. We observe that when using the mean as the aggregate, we obtain overall better performances across the different datasets and settings.

\par \noindent \textbf{Distillation.}

To improve the generalization of the embedding model, we investigate the effect of knowledge distillation by training
a new (\ie student) model using a pre-trained (\ie teacher) network with various training objectives.
\cref{tab:ablationDistill} shows a clear
performance gain with the proposed CD objective as an additional loss term, confirming the benefits of optimizing 
the learned features and relaxing the compactness of the embedding space.

Additionally, we explore sequential self-distillation similar to Born-again networks \cite{furlanello2018born},
where we consider the student model as the teacher and repeat the distillation process.
As detailed in \cref{fig:ablationDistill}, we notice a clear drop in performances beyond a single distillation step.
We suspect this might be a result of an over disentanglement of the features induced by the CD loss. As such, 
for the rest of the paper, we only apply a single distillation step to refine the features further
while preserving the learned structures. 

\par \noindent \textbf{Evaluation.}
Up until now, we primarily trained a linear classifier on top of the global features
during the meta-testing stage.
Nonetheless, given that we explicitly optimize the spatial features during training,
which increases their discriminability, we investigate their usage
as inputs to the linear classifier.
To this end, we compare the performance when training over the global features, the spatial features,
or both, where we train two classifiers and aggregate their predictions.
\cref{tab:ablationEval} shows the evaluation results. Overall, using the global features
to train the linear classifier offers slightly better results than the spatial features. 
We suspect this might result from slight overfitting of the classifier
given that the spatial features increase the number of parameters to be learned,
which negatively impacts the performances.
However, when leveraging both the spatial and global features, we obtain
better results confirming the usefulness
of the spatial feature even during the meta-testing stage.

\subsection{Few-shot Classification}

In this section, and based on the results of the ablation studies, we fix the training objective
as SC+CE during the meta-training stage and use both the spatial and global feature during
the meta-testing stage with a sum aggregate,
and compare our approach with other popular few-shot classification methods.

\par \noindent \textbf{ImageNet derivatives.}
The \textit{mini}-ImageNet benchmark is a standard dataset
used for few-shot image classification. It consists of 100 randomly selected
classes from ImageNet \cite{russakovsky2015imagenet}.
Following \cite{ravi2016optimization}, the classes are split into 64, 16, and 20
classes for meta-training, meta-validation, and meta-testing respectively.
Each class contains 600 images of size 84$\times$84.
The \textit{tiered}-ImageNet \cite{ren2018meta} benchmark presents a larger subset of ImageNet, with 608 classes
and images of size 84$\times$84
assembled into 34 super-categories. These are split into 20 categories for meta-training and 6 categories for both
meta-validation and meta-testing aiming to minimize the semantic similarity between the split.

\par \noindent \textbf{CIFAR derivatives.}
CIFAR-CS \cite{bertinetto2018meta} and FC100 \cite{oreshkin2018tadam} are both 
CIFAR-100 \cite{krizhevsky2010cifar} derivatives, containing 100 classes and images of 
size 32$\times$32. For CIFAR-CS, the classes are divided into 64, 16 and 20 classes
for meta-training, meta-validation,
and meta-testing respectively. As for FC100, the classes are grouped into 20 super-categories, which
are split into 12 categories for meta-training and 4 categories for both meta-validation and meta-testing.

\par \noindent \textbf{Results.}
The results of 5-way classification are summarized in \cref{tab:resultsImageNet} and \cref{tab:resultsCIFAR}
for ImageNet and CIFAR derivatives respectively, in addition to 20-way classification results
on \textit{mini}-ImageNet in \cref{tab:resultsMini20way}.
Our method outperforms previous works and achieves state-of-the-art performances across different
datasets and evaluation settings.
This suggests that our attention-based SCL approach coupled with the CE loss
improves the transferability of the learned
embeddings without any meta-learning techniques, with additional improvements using
a contrastive distillation step. These results also show the potential of integrating contrastive losses as
auxiliary objectives for various few-shot learning scenarios.

\subsection{Cross-Domain Few-shot Classification}

To further affirm the improved transferability of the learned embedding
with our approach,
we explore the effects of an increased domain
difference between the seen and unseen classes, \ie
the discrepancy between the meta-training and meta-testing stages. Precisely, we follow the same procedure as 
\cite{tseng2020cross} where we first train on the whole \textit{mini}-ImageNet dataset using the same setting as detailed
above. Then, we evaluate the embeddings model on four different domains: CUB \cite{welinder2010caltech},
Cars \cite{krause20133d}, Places \cite{zhou2017places}, and Plantae \cite{van2018inaturalist}.
We show the obtained results in \cref{tab:resultsCrossDomain}, and see a notable gain in performance
using the proposed method, from 2\% gain on CUB dataset, up to 7\% gain on Cars dataset, indicating a clear enhancement
in terms of the generalization of the embedding model.

\section{ProtoNet Experiments}

\begin{figure}[t]
  \centering
  \includegraphics[width=0.75\linewidth]{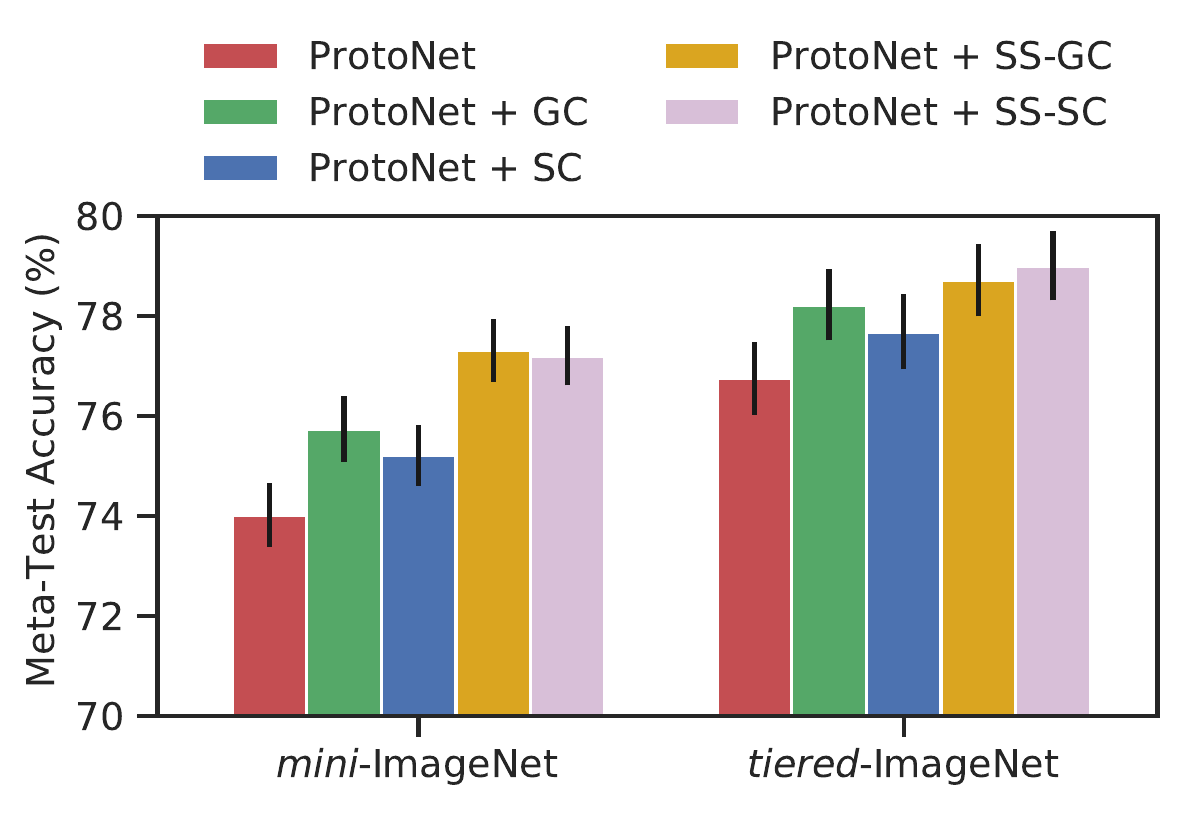}
  \vspace{-0.2in}
  \caption{The obtained improvement when adding the contrastive objectives as auxiliary losses.
  We show the mean acc. and 95\% confidence interval for 5-way 5-shot classification across ImagetNet derivatives.}
  \label{fig:ablationProto}
  \vspace{-0.05in}
\end{figure}

\begin{table}[t!]
  \setlength{\tabcolsep}{5pt}
	    \begin{center}
		\scalebox{0.71}{
		    \begin{tabular}{lcccc}
		      \toprule
		      \textbf{Method} & \textbf{Image Size}  & \textbf{Backbone} & \textbf{Aux. Loss} & \textbf{Acc. (\%)}\\

		    \midrule
		      MAML & \multirow{3}{*}{84$\times$84} & Conv4-64 & - & 63.1 \\
		      ProtoNet & & Conv4-64 & - & 68.2 \\
		      RelationNet & & Conv4-64 & - & 65.3 \\

		    \midrule
		      \multirow{2}{*}{\specialcell{ProtoNet \\ \cite{chen2019closer}}}
		      & 84$\times$84 &  Conv4-64 & - & 64.2 \\
		      & 224$\times$224 &  ResNet-18 & - & 73.7 \\
		    \midrule

		      \multirow{6}{*}{\specialcell{ProtoNet \\ \cite{gidaris2019boosting}}}
		      &\multirow{6}{*}{84$\times$84}&Conv4-64&-&70.0\\
		      &&Conv4-64&Rotation&71.7\\
		      &&Conv4-512&-&71.6\\
		      &&Conv4-512&Rotation&74.0\\
		      &&WRN-28-10&-&68.7\\
		      &&WRN-28-10&Rotation&72.1\\

		    \midrule
		      \multirow{4}{*}{\specialcell{ProtoNet \\ \cite{su2019does}}}
		      & \multirow{4}{*}{224$\times$224}&\multirow{4}{*}{ResNet-18}&-&75.2\\
		      &&&Rotation&76.0\\
		      &&&Jigsaw&76.2\\
		      &&&Rot.+Jig.&76.6\\

		    \midrule
          \cgr &\cgr&\cgr& \cgr -     & \cgr 74.0\\
		      \cgr &\cgr&\cgr& \cgr GC    & \cgr 75.2\\
		      \cgr &\cgr&\cgr& \cgr SC    & \cgr 75.2\\
		      \cgr &\cgr&\cgr& \cgr SS-GC & \cgr 77.3\\
		      \cgr &\cgr&\cgr& \cgr SS-SC & \cgr 77.2\\
          \cgr\multirow{-6}{*}{\textbf{Ours}} & \cgr\multirow{-6}{*}{\cgr224$\times$224}&\multirow{-6}{*}{\cgr ResNet-18}& \cgr SS-GC+SS-SC & \cgr \textbf{77.6}\\
		    \bottomrule

	    \end{tabular}}
    \end{center}
    \vspace{-0.15in}
    \caption{Comparison with prior works on \textit{mini}-ImageNet.
    We report the mean acc. for 5-shot 5-way classification with implementation details
    including image size, backbone model and auxiliary training losses for each method.}
    \label{tab:resultsProtonetMini}
	\vspace{-0.2in}
\end{table}

To demonstrate the generality of the proposed approach and its applicability in different settings, in this
section, we provide additional metric-learning based experiments in which we integrate the contrastive losses
into the ProtoNet \cite{snell2017prototypical} framework.
ProtoNet is a distance-based learner trained in an episodic manner, so that both the meta-training and
meta-testing stages have matching conditions.
During meta-training, for a $C$-way $K$-shot setting, we construct a meta-training set
$\cT=\{(\cD_{t}^{\mathrm{train}}, \cD_{t}^{\mathrm{test}})\}_{t=1}^{T}$ where each
given task $(\cD_{t}^{\mathrm{train}}, \cD_{t}^{\mathrm{test}})$ depicts $C$ randomly chosen classes
from the seen classes, with $K$ images per class for the training (\ie support) set $\cD_{t}^{\mathrm{train}}$,
and $M$ images per classes for the test (\ie query) set $\cD_{t}^{\mathrm{test}}$.
At each training iteration, after sampling a given task from $\cT$, we first compute the class prototypes for classification
using the support set. Then, the embeddings model is trained to minimize the CE loss where each 
query example is classified based on the distances to the class prototypes.
In order to add the contrastive objectives as auxiliary losses to the ProtoNet training objective, we simply
merge the query and support set, augment each exampled within it, and compute the contrastive losses detailed in \cref{sec:methods}
over this merged and augmented set.

\par \noindent \textbf{Experimental Details.} For the experimentation, we follow \cite{chen2019closer}
and base our implementation on their few-shot learning code base. In particular, we use a ResNet-18
network as the embedding model with 512-dimensional output features. We train on ImagetNet derivatives using ADAM optimizer
with a learning rate of $10^{-3}$ for 60,000 episodes
and use 5-way (classes) 5-shot (examples per-class) with 16 query images.
For contrastive learning, similar to $\cref{sec:exp}$, we use and a two-layer MLP for the projection head
and the attention modules with an output dimensionality of $64$, and set $\lambda_{\mathrm{CE}}$ to $1.0$,
and $\lambda_{\mathrm{GC}}$ and $\lambda_{\mathrm{SC}}$ to $0.5$.
For meta-testing, we report the mean accuracy and 95\% confidence interval over 600 randomly sampled
tasks, where each class consists of 5 support images and 16 query images.

\par \noindent \textbf{Results.} To investigate the impact of the contrastive
losses on the performances of ProtoNet, we report the mean acc. for 5-way 5-shot classification on ImageNet
derivatives with different training objectives.
The results in \cref{fig:ablationProto} show a notable performance gain over the ProtoNet baseline.
Surprisingly, when disregarding the labels and training with the self-supervised formulation of the
contrastive objectives, we obtain better results. The SS-SC and SS-GC losses perform comparatively
on \textit{mini}-ImageNet with a 3.2\% gain, and with the SS-SC loss performing slightly better
on \textit{tiered}-ImageNet with a 2.2\% gain.
We suspect that the obtained gain when using the self-supervised formulation might be
a result of using a larger number of negatives as opposed to the supervised formulation,
since each batch of examples only contains $5$ unique classes.
Additionally, we compare the performances of our approach with other self-supervised auxiliary losses,
\ie rotation prediction \cite{gidaris2018unsupervised} and jigsaw puzzle \cite{noroozi2016unsupervised}, for which
\cite{su2019does} provided their integration into the ProtoNet framework.
As shown in \cref{tab:resultsProtonetMini}, we observe that a larger performance gain can
be obtained with the contrastive objectives as auxiliary losses compared
to other self-supervised objectives, especially when using both the SS-SC and SS-GC losses
with a 3.6\% gain over the baseline, which further confirms the effectiveness of the proposed SC loss.

\section{Related Work}
\label{sec:rw}

\par \noindent \textbf{Few-Shot Classification.}
In few-shot classification, the objective is to learn to recognize unseen novel classes
with few labeled example in each class. Meta-learning remains the most popular paradigm
to tackle this problem. Roughly, such approaches can be divided into two categories.
Optimization-based, or \textit{learning to learn} methods
\cite{ravi2016optimization, andrychowicz2016learning, wang2016learning, finn2017model, sun2019meta,
lee2019meta, rusu2018meta}, that integrate the fine-tuning process in the meta-training algorithm to
rapidly adapt to model to the unseen classes with limited supervision.
And metric-based, or \textit{learning to compare} methods
\cite{vinyals2016matching, snell2017prototypical, sung2018learning, oreshkin2018tadam, scott2018adapted, doersch2020crosstransformers},
that learn a common embedding space in which the similarities between
the data can help distinguish between different novel categories with a given distance metric.
Most relevant to our work are the methods that follow the standard transfer learning strategy
\cite{chen2019closer,dhillon2019baseline,tian2020rethinking,afrasiyabi2019associative}.
Consisting of two stages, a pre-training stage with the CE loss on the meta-training set,
then a fine-tuning stage on the meta-testing set.
Despite their apparent simplicity, \cite{tian2020rethinking} showed that such a strategy yields
state-of-the-art results on standard benchmarks.

\par \noindent \textbf{Cross-Entropy Loss.}
The CE loss continues to be the prominent supervised learning objective
used for training deep networks, in which the model is trained to predict the
corresponding class label in the form of a one-hot vector.
However, despite its success, some works showed many possible drawbacks \cite{khosla2020supervised},
\eg noise sensitivity \cite{sukhbaatar2014training,zhang2018generalized}, adversarial examples \cite{nar2019cross},
and suboptimal margins \cite{elsayed2018large,cao2019learning}.
While other works proposed some alternative approaches, such as changing the label distribution
\cite{szegedy2016rethinking,yun2019cutmix,muller2019does,zhang2017mixup}
or leveraging the contrastive losses \cite{khosla2020supervised}.

\par \noindent \textbf{Contrastive Learning.}
Instead of training the network to match to the input to a fixed target.
Contrastive learning acts directly on the low-dimensional representations with
contrastive losses \cite{hadsell2006dimensionality,gutmann2010noise, salakhutdinov2007learning},
that measure the similarities of different samples in the embedding space.
Recently, contrastive learning based methods have emerged as the state-of-the-art
approaches for self-supervised representation learning. The main difference
between them is the way they construct and choose the positive samples.
In this work, we differentiate between self-supervised contrastive methods
\cite{oord2018representation, wu2018unsupervised, hjelm2018learning, henaff2019data,
tian2019contrastive, he2020momentum, chen2020simple}
that leverage data augmentations to construct the positive pairs, and supervised contrastive methods
\cite{salakhutdinov2007learning,wu2018improving,kamnitsas2018semi,khosla2020supervised}
that leverage the provided labels to sample the positive examples.

\par \noindent \textbf{Self-Supervised Learning and Few-Shot Classification.}
Relevant to our work are methods that try to 
build on the insights and advances in contrastive learning, or more broadly self-supervised learning,
to improve the few-shot classification task. Such methods
\cite{gidaris2019boosting,medina2020self,su2019does,doersch2020crosstransformers,gao2021contrastive}
integrate various types of self-supervised training objectives into different few-shot learning frameworks in order to learn more transferable features
and improve the few-shot classification performance. In this paper,
we propose a novel contrastive learning objective based on the spatial features
to further promote general purpose and robust representations suited for few-shot classification.
In this context, a similar idea was proposed in \cite{doersch2020crosstransformers}.
In their approach, a constrastive pre-training stage is first
conducted followed by the standard ProtoNet \cite{snell2017prototypical} 
fine-tuning stage where spatial features are used to compute the similarity between the
training and testing instances.
In our work, contrary to \cite{doersch2020crosstransformers}, we integrate the spatial information directly into the contrastive learning loss.  The proposed loss is then integrated into the training as an auxiliary loss,
resulting in a far more effective, flexible and general framework
usable in various few-shot learning scenarios.

\section*{Conclusion}
In this paper, we investigated contrastive losses as auxiliary training objectives along the CE loss to compensate for its drawbacks and learn richer and more transferable features.
With extensive experiments, we showed that integrating contrastive learning into existing few-shot learning frameworks results in a notable boost in performances,
especially with our spatial contrastive learning objective.
Future work could investigate the spatial contrastive method extension for
other few-shot learning scenarios and adapt it for other visual tasks such as unsupervised representation learning.

\section*{Acknowledgements}
The first author is supported by Randstad corporate research chair,
in collaboration with CentraleSupélec, Université Paris-Saclay.
This work was performed using HPC resources from the Mésocentre computing center
of CentraleSupélec and
École NormaleSupérieure Paris-Saclay supported by CNRS and Région Île-de-France
(\footnotesize{\url{http://mesocentre.centralesupelec.fr/}}).

\bibliography{paper}
\bibliographystyle{icml2020}

\cleardoublepage
\appendix
\onecolumn
\counterwithin{figure}{section}
\counterwithin{table}{section}

\normalsize

\section*{\hspace{2.5in} Supplementary Material}

\section{Datasets}

In this paper, we experimented with different datasets used for various few-shot classification settings.
For the standard few-shot classification setting, we used the four popular benchmarks: \textit{mini}-ImageNet, 
\textit{tiered}-ImageNet, CIFAR-FS and FC100. As for cross-domain few-shot classification, following \cite{tseng2020cross},
we train on the whole \textit{mini}-ImageNet (\ie train, val and test sets), and then we evaluate on one of the following datasets:
CUB, Cars, Places or Plantae. The details about each dataset are presented in \cref{tab:appendixDatasets}.

\begin{table*}[!ht]
\setlength{\tabcolsep}{0.4cm}
	\small
	\centering
	\scalebox{0.8}{
	\begin{tabular}{l c c c c c}
	\toprule
	Datasets & Source & Nbr. of train classes & Nbr. of val classes & Nbr. of test classes & Split setting \\
	\midrule
	\textit{mini}-ImageNet & \cite{vinyals2016matching} & 64 & 16 & 20 & \cite{ravi2016optimization} \\ 
	\textit{tiered}-ImageNet & \cite{ren2018meta} & 351 & 97 & 160 & Original \\ 
	CIFAR-FS & \cite{bertinetto2018meta} & 64 & 16 & 20 & Original \\ 
	FC100 & \cite{oreshkin2018tadam} & 60 & 20 & 20 & Original \\ 
	CUB & \cite{welinder2010caltech} & 100 & 50 & 50 & \cite{hilliard2018few} \\ 
	Cars & \cite{krause20133d} & 98 & 49 & 49 & Random \\ 
	Places & \cite{zhou2017places} & 183 & 91 & 91 & Random \\ 
	Plantae & \cite{van2018inaturalist} & 100 & 50 & 50 & Random \\ 
    \bottomrule
	\end{tabular}}
	\vspace{-0.1in}
	\caption{
	Datasets. Additional details about the datasets used in the experiments.}
	\vspace{-0.1in}
	\label{tab:appendixDatasets}
\end{table*}

\section{Additional Experiments}

\begin{table*}[!t]
\setlength{\tabcolsep}{0.4cm}
	\small
	\centering
	\scalebox{0.8}{
		\begin{tabular}{l l c c c c c c}
		\toprule
		&  & \multicolumn{2}{c}{\textit{mini}-ImageNet, 5-way} & \multicolumn{2}{c}{CIFAR-FS, 5-way} & \multicolumn{2}{c}{FC100, 5-way} \\
		\multirow{-2}{*}{Method} & \multirow{-2}{*}{Backbone} & 1-shot & 5-shot &  1-shot & 5-shot &  1-shot & 5-shot \\
	\midrule
    Ours                   & 64-64-64-64     &  50.45 $\pm$ 0.83    & 67.88 $\pm$ 0.61    & 60.31 $\pm$ 0.92   & 77.15 $\pm$ 0.66 & 38.45 $\pm$ 0.72 & 52.15 $\pm$ 0.77 \\
    Ours-Distill           & 64-64-64-64     &  52.18 $\pm$ 0.89    & 68.45 $\pm$ 0.68    & 60.8 $\pm$ 0.88    & 77.25 $\pm$ 0.67 & 38.31 $\pm$ 0.75 & 52.06 $\pm$ 0.77 \\

    Ours-Trainval          & 64-64-64-64     &  51.74 $\pm$ 0.76    & 68.69 $\pm$ 0.74    & 60.75 $\pm$ 0.92   & 77.56 $\pm$ 0.72 & \bt{42.07 $\pm$ 0.74} &  55.97 $\pm$ 0.73 \\
    Ours-Distill-Trainval  & 64-64-64-64     &  \bt{52.64 $\pm$ 0.84}    & \bt{69.46 $\pm$ 0.61}    & \bt{62.02 $\pm$ 0.93}   & \bt{77.88 $\pm$ 0.76} & 41.73 $\pm$ 0.75 & \bt{56.16 $\pm$ 0.75} \\
	\midrule
    Ours                   & ResNet-12      &  65.66 $\pm$ 0.76 &  82.52 $\pm$ 0.50 & 75.01 $\pm$ 0.91 & 87.44 $\pm$ 0.58 & 44.30 $\pm$ 0.70   & 59.80 $\pm$ 0.70 \\
    Ours-Distill           & ResNet-12      &  67.40 $\pm$ 0.76 & 82.70 $\pm$ 0.52 & 76.46 $\pm$ 0.87 & 87.62 $\pm$ 0.59 & 44.80 $\pm$ 0.70   & 61.40 $\pm$ 0.70 \\

    Ours-Trainval          & ResNet-12      &  67.02 $\pm$ 0.81    & 84.01 $\pm$ 0.53    & 76.08 $\pm$ 0.87   & 87.69 $\pm$ 0.62 & 50.89 $\pm$ 0.79 & 67.57 $\pm$ 0.75\\
    Ours-Distill-Trainval  & ResNet-12      &  \bt{68.54 $\pm$ 0.84}    & \bt{84.50 $\pm$ 0.50}    & \bt{77.88 $\pm$ 0.81}   & \bt{88.57 $\pm$ 0.61} & \bt{51.12 $\pm$ 0.81} & \bt{67.96 $\pm$ 0.69} \\
	\midrule
    Ours                   & SEResNet-12     &  65.83 $\pm$ 0.76    & 81.66 $\pm$ 0.58    & 74.70 $\pm$ 0.92   & 86.90 $\pm$ 0.59 & 42.45 $\pm$ 0.71 & 59.72 $\pm$ 0.71 \\
    Ours-Distill           & SEResNet-12     &  66.38 $\pm$ 0.81    & 83.25 $\pm$ 0.50    & 76.41 $\pm$ 0.88   & 87.44 $\pm$ 0.62 & 43.00 $\pm$ 0.73 & 60.84 $\pm$ 0.76 \\

    Ours-Trainval          & SEResNet-12     &  67.84 $\pm$ 0.72    & 83.38 $\pm$ 0.55   & 75.54 $\pm$ 0.87   & 87.27 $\pm$ 0.60 & \bt{51.70 $\pm$ 0.78} & 68.19 $\pm$ 0.73 \\
    Ours-Distill-Trainval  & SEResNet-12     &  \bt{69.28 $\pm$ 0.84}    & \bt{83.92 $\pm$ 0.52}    & \bt{76.72 $\pm$ 0.82}   & \bt{87.86 $\pm$ 0.59} & 51.06 $\pm$ 0.83 & \bt{68.28 $\pm$ 0.73} \\
    \bottomrule
	\end{tabular}}
	\caption{Comparison of different backbones on 5-way classification on \textit{mini}-ImageNet, CIFAR-FS and FC100.
	We show the mean acc. and 95\% confidence interval.}
	\label{tab:appendixModel}
\end{table*}

\begin{table*}[!t]
\setlength{\tabcolsep}{0.4cm}
	\small
	\centering
	\scalebox{0.8}{
		\begin{tabular}{l c c c c c c c}
		\toprule
		& & \multicolumn{2}{c}{\textit{mini}-ImageNet, 5-way} & \multicolumn{2}{c}{CIFAR-FS, 5-way} & \multicolumn{2}{c}{FC100, 5-way} \\
		\multirow{-2}{*}{Features Used} & Weight Imp. & 1-shot & 5-shot &  1-shot & 5-shot &  1-shot & 5-shot \\
	\midrule
    Spatial                   & &  63.40 $\pm$ 0.86    & 81.42 $\pm$ 0.52    & 74.56 $\pm$ 0.89   & 86.83 $\pm$ 0.64 & 42.92 $\pm$ 0.71 & 59.49 $\pm$ 0.74 \\
    Global                    & &  64.49 $\pm$ 0.85    & 81.87 $\pm$ 0.58    & 74.96 $\pm$ 0.89   & 87.65 $\pm$ 0.59 & 43.90 $\pm$ 0.77 & 60.61 $\pm$ 0.76 \\
    Glo. \& Spa. (Max)        & &  64.94 $\pm$ 0.78    & 82.18 $\pm$ 0.50    & 74.40 $\pm$ 0.87   & 87.18 $\pm$ 0.58 & \bt{43.56 $\pm$ 0.75} & \bt{60.78 $\pm$ 0.71} \\
    Glo. \& Spa. (Sum)        & &  65.30 $\pm$ 0.77    & 81.53 $\pm$ 0.55    & 74.30 $\pm$ 0.85   & \bt{87.72 $\pm$ 0.58} & 43.18 $\pm$ 0.73 & 60.11 $\pm$ 0.70 \\
	\midrule
    Spatial                   & $\checkmark$ &  64.12 $\pm$ 0.82    & 81.98 $\pm$ 0.53    & 74.93 $\pm$ 0.81   & 87.30 $\pm$ 0.59 & 43.87 $\pm$ 0.71 & 60.64 $\pm$ 0.82 \\
    Global                    & $\checkmark$ &  65.14 $\pm$ 0.81    & \bt{82.72 $\pm$ 0.58}    & \bt{75.43 $\pm$ 0.88}   & 87.18 $\pm$ 0.56 & 43.21 $\pm$ 0.77 & 59.53 $\pm$ 0.72 \\
    Glo. \& Spa. (Max)        & $\checkmark$ &  64.91 $\pm$ 0.79    & 81.60 $\pm$ 0.57    & 74.39 $\pm$ 0.84   & 87.22 $\pm$ 0.60 & 43.54 $\pm$ 0.72 & 59.44 $\pm$ 0.77 \\
    Glo. \& Spa. (Sum)        & $\checkmark$ &  \bt{65.56 $\pm$ 0.84}    & 82.43 $\pm$ 0.54    & 75.08 $\pm$ 0.87   & 87.30 $\pm$ 0.57 & \bt{43.56 $\pm$ 0.68} & 60.04 $\pm$ 0.70 \\
    \bottomrule
	\end{tabular}}
	\caption{
	Comparison of different evaluation setups. We train a linear classifier on
	either the spatial, the global features, or both, where
	two classifiers are used and their predictions are aggregated by either
	taking their sum or maximum per class. We also investigate
	the effect of weight imprinting to improve the initialization
	of the classifier at test time.
	Note that the presented results use a PyTorch implementation of
	the classifiers instead of scikit-learn, resulting in slight decrease in performances.}
	\label{tab:appendixEval}
\end{table*}

\begin{table*}[!t]
\setlength{\tabcolsep}{0.4cm}
	\small
	\centering
	\scalebox{0.8}{
		\begin{tabular}{l@{\hskip 1in} c c c c c c}
		\toprule
		& \multicolumn{2}{c}{\textit{mini}-ImageNet, 5-way} & \multicolumn{2}{c}{CIFAR-FS, 5-way} & \multicolumn{2}{c}{FC100, 5-way} \\
		\multirow{-2}{*}{Contrastive Distillation} & 1-shot & 5-shot &  1-shot & 5-shot &  1-shot & 5-shot \\
	\midrule
    \textit{Teacher} & \textit{65.7 $\pm$ 0.8} &  \textit{82.5 $\pm$ 0.5} & \textit{75.0 $\pm$ 0.9} & \textit{87.4 $\pm$ 0.6} & \textit{44.4 $\pm$ 0.8} & \textit{60.8 $\pm$ 0.9} \\
    \addlinespace[0.05in]
    Global ($\beta = 0$)          &   \bt{67.4 $\pm$ 0.8}    & \bt{83.2 $\pm$ 0.5}    & \bt{76.5 $\pm$ 0.9}    & \bt{88.0 $\pm$ 0.6}    & \bt{44.8 $\pm$ 0.7}    & \bt{61.4 $\pm$ 0.7} \\
    Spatial ($\alpha = 0$)        &   66.0 $\pm$ 0.8    & \bt{83.2 $\pm$ 0.5}    & 76.0 $\pm$ 0.8    & \bt{80.0 $\pm$ 0.5}    & 44.4 $\pm$ 0.7    & 61.0 $\pm$ 0.7 \\
    Global \& Spatial             &   65.6 $\pm$ 0.8    & 83.1 $\pm$ 0.5    & 75.4 $\pm$ 0.9    & 87.3 $\pm$ 0.6    & 44.3 $\pm$ 0.8    & 61.2 $\pm$ 0.8 \\
	\bottomrule
	\end{tabular}}
	\caption{
	Comparison of contrastive distillation losses on 5-way classification on \textit{mini}-ImageNet, CIFAR-FS and FC100.
	We show the mean acc. and 95\% confidence interval.}
	\label{tab:appendixDistill}
\end{table*}

\subsection{Embedding Model}
\begin{figure}[!ht]
  \centering
  \includegraphics[width=0.6\linewidth]{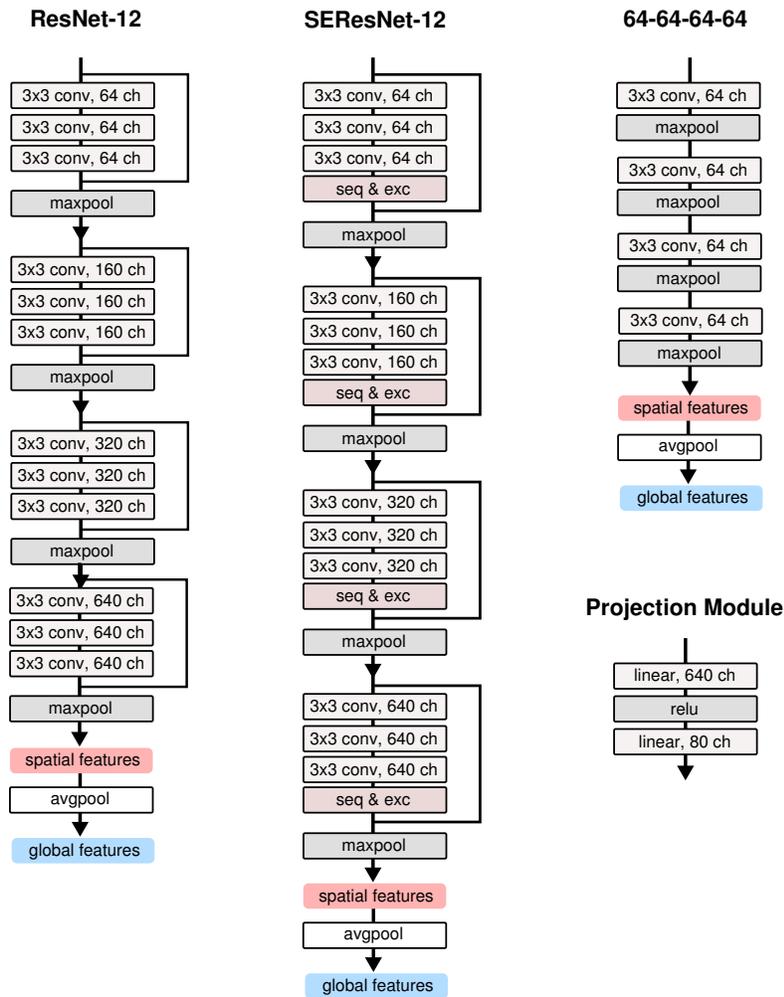}
  \caption{Architectural details. The architecture of the backbones used as embeddings models, and the projection module used
  both as the projection head for GC loss, and for the attention modules (\ie head, query and value heads) for the SC loss.
  ``seq \& exc'' refers to squeeze and excitation modules.}
  \label{fig:Appendixmodels}
\end{figure}

In the transfer learning experiments, we mainly used a ResNet-12 backbone
as the embedding model. Since the backbone also
has a significant impact on the quality of the produced embedding,
we experiment with various backbones. In particular, we compare 
ResNet-12 to two other alternative: 4 layer convolution network
(\ie 64-64-64-64) and ResNet-12 with sequeeze-and-excitation \cite{hu2018squeeze}
layers that play the role of a channel-wise attention module.
See \cref{fig:Appendixmodels} for an illustration of these backbones.
The results are presented in \cref{tab:appendixModel},
where we train the models either on the meta-training set only
or on both the meta-training and the meta-validation set. In both cases with and without a contrastive distillation step.
As expected, we observe that, (1) better models improve the
performances, and (2) more training data yield better results, further 
emphasizing the importance of learning a well performing and transferable embedding
model for an effective few-shot classification.

\subsection{Weight Imprinting}

During the ablation experiments, and when training a base classifier on top
of the spatial features, a case that requires a larger classifier, we observed
a slight decrease in the performances, which we suspect might be due to the
overfitting of the model, especially at low shot setting. To overcome this, we
investigate using imprinted weights \cite{qi2018low} to initialize the base
classifier and help stabilize the convergence.
Weight imprinting consists of directly setting the base classifier's weights
from novel training examples during low-shot learning \ie the $\ell_2$ normalized class
prototypes constructed from the training (\ie support) features.
This process is called weight imprinting 
since it directly sets the weights of the base classifier based on an the
scaled features produced by the embedding model for a given training example at test time.
\cref{tab:appendixEval} summarizes the results, although the weight imprinting
does slightly help the performances, the results are not conclusive and need further investigation.

\subsection{Contrastive Distillation}

As detailed in the paper, in order to relax the placement of the embeddings
when applying a distillation step, we also optimize a contrastive loss $L_{\mathrm{CD}}$ loss in addition to the 
standard KL loss. The loss consists of aligning the global features of the teacher and the student
by reducing their cosine similarity. However, we can also align the spatial
features rather than the global features, or both. In this case, the contrastive loss is written as follows:

\begin{equation}
L_{\mathrm{CD}} = \alpha \frac{1}{N} \sum_{i=1}^{N} \|\mbz^{\rmg t}_i - \mbz^{\rmg s}_i \|_{2}^{2} +
\beta \frac{1}{N} \sum_{i=1}^{N} \|\mbz^{\rms t}_i - \mbz^{\rms s}_i \|_{2}^{2}.
\end{equation}

The results are shown in \cref{tab:appendixDistill}. We observe while contrasting only the global
features works slightly better, the obtained results are overall similar.

\section{Additional Experimental Details}

\textbf{Training.}
As stated in the paper, during training, we reduce the learning rate with a
factor of 0.1 at different training iterations.
For \textit{mini}-ImageNet, we train for 90 epochs and we decay the learning
rate at 60 and 80 epochs, for \textit{tiered}-ImageNet,
we train for 60 epochs and decay the learning rate
rate three times, at 30, 40 and 50 epochs. As for
CIFAR-100 derivatives, we train for 90 epochs and decay the learning
rate three times at 45, 60 and 75 epochs for CIFAR-FS, while for
FC-100, we train for 65 epochs with single learning decay step at 60 epochs.
For distillation, we change to learning rate to $10^{-2}$ and train with similar settings.

For the weights of the loss functions, while the weight of the cross-entropy loss is always set to 1, 
when training with only one contrastive loss as an auxiliary loss, we set its weight to 1,
\ie $\lambda_{\mathrm{GC}} = \lambda_{\mathrm{SC}} = 1$, 
be it the supervised of self-supervised formulations,
except for CIFAR-FS where we set $\lambda_{\mathrm{GC}} = \lambda_{\mathrm{SC}} = 0.5$.
Additionally, when training with both objectives,
we set $\lambda_{\mathrm{SC}} = \lambda_{\mathrm{GC}} = 0.5$. 

For transfer learning experiments, we train on either 84$\times$84 sized images for ImageNet derivatives,
resulting in spatial features of spatial dimensions of 5$\times$5, or
32$\times$32 sized images for CIFAR-100 derivatives, resulting in spatial features of spatial dimensions of 2$\times$2.
For ProtoNet experiments, we train on 224$\times$224 sized images with ResNet-18 backbone, resulting in spatial features of 
spatial dimensions of 7$\times$7. Thus, in order to reduce the computational
requirement when applying the attention based spatial alignment,
we apply an adaptive average pooling, reducing the dimensions to 3$\times$3
instead of 7$\times$7 resulting in both better results and faster training time.

\textbf{Evaluation.} During evaluation, we use a multivariate logistic regression implemented in
scikit-learn \cite{pedregosa2011scikit} trained on the $\ell_2$ normalized
global features. Additionally, when using the spatial features during evaluation, and in order to
reduce the number of parameters of the linear classifier and avoid overfitting, we first
apply a max pooling operation over these features, reducing their spatial dimensions
to $2 \times 2$, and then feed them to the linear classifier.
As for cross-domain experiments, we found that augmenting each training (\ie support)
samples 10 times instead of 5 produces slightly better results. 

When conducting ablation studies with weight imprinting (\ie \cref{tab:appendixEval}),
we use our own implementation of the
multivariate logistic regression with an L-BFGS optimizer and an $\ell_2$ penalty.
While the performances are overall similar, this gives us more degrees of freedom when implementing the base
classifier. Specifically, when using the spatial features as input to the base classifier,
we implement the classifier as a convolutional layer, in which the filter
size matches the dimensions of the spatial features.

\section{Architectures}

\section{Quality of the Learned Representations}

\begin{figure}[!ht]
  \centering
  \includegraphics[width=0.98\linewidth]{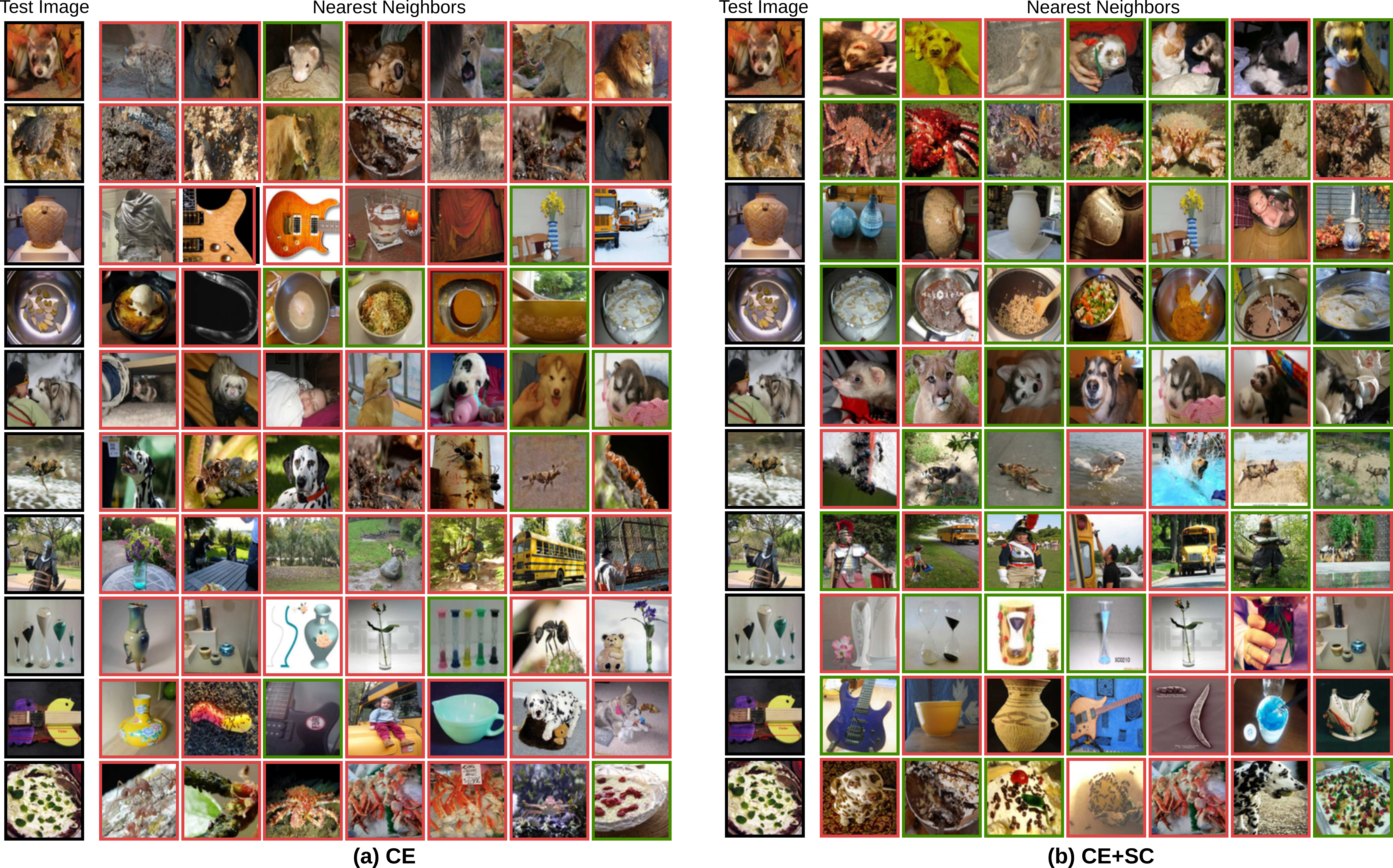}
  \caption{{Nearest Neighbors Analysis.} 
  For a given test image from \textit{mini}-ImageNet dataset,
  we compute the nearest neighbors in the embedding space on the test set
  when a model is train with either, (a) the standard CE, or, (b) with the proposed SC
  as an additional auxiliary objective.
  We observe that the neighboring images in the embedding space
  found when the SC loss is used are more semantically similar then
  the standard case with the CE loss.
  It suggests that the quality of the learned embeddings is increased with
  the usage of the SC as an auxiliary loss as a result of optimizing for more general-purpose features.}
  \label{fig:AppendixFeaturesAnalysis}
\end{figure}

\begin{figure}[!ht]
  \centering
  \includegraphics[width=0.52\linewidth]{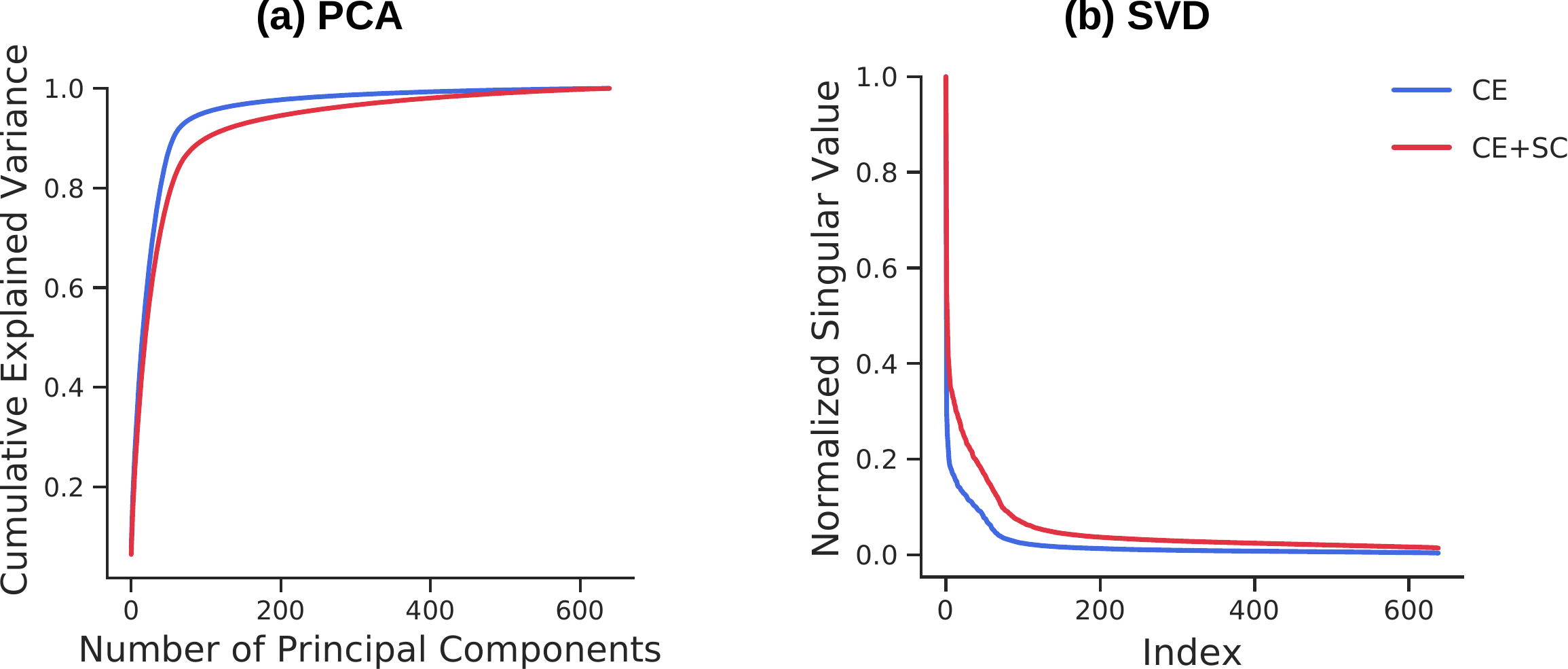}
  \caption{{Spectral Analysis.} 
  Results of the spectral analysis
  on the embedding matrix using CE or CE+SC as training objectives.
  The plot (a) shows the explained
  cumulative variance of the learned features as the number of principal components
  used and (b) shows the max-normalized singular values. We observe that the SC loss
  increases the number of dominant principal components and the weight
  of the remaining singular values, which indicates that the SC loss
  does help retain more informative signals that might be useful outside of the meta-training classification task.}
  \label{fig:AppendixSpectralAnalysis}
\end{figure}

In this section, and similar to the analysis conducted in the paper, we conduct
an empirical analysis of the embeddings to assess the quality of the learned features in two cases: (1)
when the model is trained with only the CE loss, and (2) when adding the SC as
an additional auxiliary objective. \cref{fig:AppendixFeaturesAnalysis,fig:AppendixSpectralAnalysis} show the results. We observe a clear improvement
in terms of the obtained nearest neighbors when using the SC loss and
also an increase in terms of the amount of informative signal retrained within 
the embedding matrix, both indicating an enhancement in the quality of the learned embeddings.

\end{document}